\documentclass[lettersize,journal]{IEEEtran}
\usepackage{amsmath,amsfonts}
\usepackage{algorithmic}
\usepackage{algorithm}
\usepackage{array}
\usepackage[caption=false,font=normalsize,labelfont=sf,textfont=sf]{subfig}
\usepackage{textcomp}
\usepackage{stfloats}
\usepackage{url}
\usepackage{verbatim}
\usepackage{graphicx}

\usepackage{amssymb}
\usepackage{booktabs}
\usepackage{multirow}
\usepackage{makecell}
\usepackage{array}
\usepackage{xcolor}
\usepackage{xspace}
\usepackage{colortbl}

\usepackage[colorlinks]{hyperref}
\usepackage[capitalize]{cleveref}

\crefname{section}{Sec.}{Secs.}
\Crefname{section}{Section}{Sections}
\Crefname{table}{Table}{Tables}
\crefname{table}{Tab.}{Tabs.}

\makeatletter
\DeclareRobustCommand\onedot{\futurelet\@let@token\@onedot}
\def\@onedot{\ifx\@let@token.\else.\null\fi\xspace}

\def\eg{\emph{e.g}\onedot} 
\def\ie{\emph{i.e}\onedot}

\def\etal{\emph{et al}\onedot}
\makeatother

\usepackage{tikz}

\newcommand\copyrighttext{%
  \footnotesize \textcolor{red}{\textbf{\textcopyright\,2024 IEEE. Personal use of this material is permitted. Permission from IEEE must be obtained for all other uses, in any current or future media, including reprinting/republishing this material for advertising or promotional purposes, creating new collective works, for resale or redistribution to servers or lists, or reuse of any copyrighted component of this work in other works.}}}
\newcommand\copyrightnotice{%
\begin{tikzpicture}[remember picture,overlay]
\node[anchor=south,yshift=3pt] at (current page.south) {\fbox{\parbox{\dimexpr\textwidth-\fboxsep-\fboxrule\relax}{\copyrighttext}}};
\end{tikzpicture}%
}

\begin{document}
\bstctlcite{IEEEexample:BSTcontrol}

\title{One-shot Human Motion Transfer via Occlusion-Robust Flow Prediction and Neural Texturing}

\author{Yuzhu~Ji*,~Chuanxia~Zheng,~and~Tat-Jen~Cham
\thanks{Yuzhu Ji is with the School of Computer Science and Technology, Guangdong University of Technology, Guangzhou 510006, P. R. China; Chuanxia~Zheng is with the Department of Engineering Science, University of Oxford, UK;
Tat-Jen~Cham is with the College of Computing and Data Science, Nanyang Technological University, Singapore. \emph{(Corresponding author: Yuzhu Ji, E-mail: yuzhu.ji@gdut.edu.cn.)}
}
}

\markboth{This article has been accepted for publication in IEEE Transactions on Multimedia. DOI:XX.XXX/XXX.XX.XXXX}%
{Shell \MakeLowercase{\textit{et al.}}: A Sample Article Using IEEEtran.cls for IEEE Journals}


\newcommand{\YZ}[1]{{\color{red}{YZ: #1}}}

\maketitle

\copyrightnotice

\begin{abstract}
Human motion transfer aims at animating a static source image with a driving video. While recent advances in one-shot human motion transfer have led to significant improvement in results, it remains challenging for methods with 2D body landmarks, skeleton and semantic mask to accurately capture correspondences between source and driving poses due to the large variation in motion and articulation complexity. In addition, the accuracy and precision of DensePose degrade the image quality for neural-rendering-based methods. To address the limitations and by both considering the importance of appearance and geometry for motion transfer, in this work, we proposed a unified framework that combines multi-scale feature warping and neural texture mapping to recover better 2D appearance and 2.5D geometry, partly by exploiting the information from DensePose, yet adapting to its inherent limited accuracy. Our model takes advantage of multiple modalities by jointly training and fusing them, which allows it to robust neural texture features that cope with geometric errors as well as multi-scale dense motion flow that better preserves appearance. Experimental results with full and half-view body video datasets demonstrate that our model can generalize well and achieve competitive results, and that it is particularly effective in handling challenging cases such as those with substantial self-occlusions.

\end{abstract}


\section{Introduction}
\label{sec:intro}
\IEEEPARstart{H}{uman} motion transfer refers to the task of animating a static image of a person to fit the corresponding \emph{motion} in a driving video sequence while preserving the original person's \emph{identity}; thus learning to disentangle the \emph{motion} and \emph{identity} is of central importance. However, due to the diversity of motion patterns, different identities along with different body shapes, and self-occlusion issues, it remains challenging to render photo-realistic human images based on specified driving motions and generate controllable human animations \cite{liu2019neural,cgfTewariFTSLSMSSN20}. 

Recent advances in motion transfer have adopted two main paradigms: 1) pose-guided image/video generation and 2) neural-rendering-based methods. In particular, the pose-guided image/video generation methods directly warp appearance features through various correspondences between source and driving pose, including through dense motion flow~\cite{cvprSiarohinSLS18,nipsSiarohinLT0S19,cvprSiarohinWRCT21,TaoG0D23flow}, attention~\cite{cvprZhuHSYWB19,RenF0LLcvpr22,cvprBhuniaKCALSK23} and cross-domain correlations~\cite{Zhang_2020_CVPR,Zhou_2021_CVPR}. These methods can achieve good results in preserving identity, but they do not handle large variations in body pose well. A key issue is the \emph{self-occlusion} within the motion that corrupts the accuracy of correspondence for feature alignment, resulting in reduced quality of synthesized images. Besides, such methods may be confused about whether a person is facing front or back when using degenerate 2D supervision, \eg 2D landmarks, articulated structures, and semantic masks. 

\begin{figure}
  \centering
  \includegraphics[trim = 0mm 0mm 0mm 0mm, clip=true, width=0.9\linewidth]{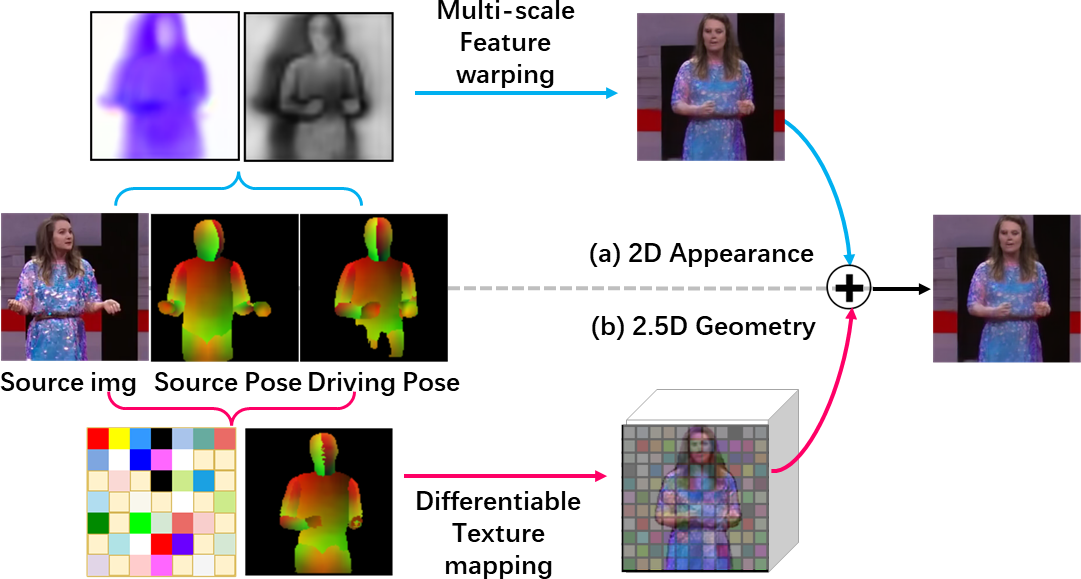}
  \caption{\textbf{Broad view of our approach} for human motion transfer allows for good fidelity in 2D appearance transfer by estimating 2D motion flow, while establishing pose accuracy through 2.5D geometric reasoning. See our comparative results in Fig.~\ref{fig2:cvanimation}.}
  \label{fig0:teaserimg}
\end{figure}

Separately, neural-rendering-based methods can achieve high-fidelity and photo-realistic human image generation and animation~\cite{nipsWang0ZYTKC18,iccvChanGZE19,nipsWang0TLCK19}. However, most of these models need to be trained in a person-specific fashion and can \emph{not easily be generalized to new individuals}~\cite{cgfTewariFTSLSMSSN20,SunHWLLGpami23}. Recently, efforts~\cite{eccvNeverovaGK18,eccvSarkarMXGT20,huang2021few,tcsvtYangYZJZS22,tmmLiuZCZ22, corrdreampose23} have been made to generalize neural-rendering-based methods to arbitrary persons by utilizing auxiliary 2.5D or 3D prior information, such as DensePose IUV map \cite{cvprGulerNK18} and the Skinned Multi-Person Linear (SMPL) model \cite{togLoperM0PB15}. {Specifically, the DensePose IUV map defines the SMPL silhouette of the human body~\cite{cvprGulerNK18}, and the UV coordinate system of the DensePose is a parameterized representation of the SMPL model's surface. In this way, the DensePose establishes a 2.5D dense correspondence between the 2D image and the 3D human model. However, these auxiliary structure information can be noisy, which may result in a significant loss of 2D appearance information. In particular, the DensePose IUV map may occasionally suffer from large gaps. SMPL \cite{togLoperM0PB15} is a 3D model that only models the human body without hair and clothing. In practice, the parameters of SMPL model can be significantly influenced by occlusions.}

The limitations of these two main paradigms motivate us to introduce a novel architecture that can jointly learn the \emph{appearance} and \emph{geometry} in a unified pipeline (see Fig.~\ref{fig0:teaserimg}). Here, our \emph{key insight} is that despite its limited accuracy and restriction to the SMPL silhouette, the DensePose IUV representation can be used as guidance in two ways: (a) it can be used to predict a more accurate dense motion flow for feature warping by combining with the \emph{appearance} context of the source image, even for regions that are \emph{not} originally covered by DensePose; and (b) it can be used to support texture atlas translation and mapping, via neural textures that can overcome inaccuracies in coarse geometry. Furthermore, the joint training of 2.5D geometry and 2D appearance features may correct geometrical errors and stabilize the generation of images through multi-modal feature fusion.

To achieve this goal, we present a unified framework that learns to synthesize images with better recovery of \emph{appearance} and \emph{geometry} by combining the advantages of pose-guided and neural-rendering-based methods. {The framework aims to leverage 2.5D DensePose IUV maps to improve geometry recovery while accounting for their inherent limited accuracy. In particular, we employ a 2D branch network with motion flow warping to enhance appearance recovery, while a 2.5D branch network focuses on recovering geometry. Though DensePose estimates 2.5D geometry in the target, the available pre-trained model struggled to generate accurate IUV maps in challenging occlusion scenarios. Therefore, the 2.5D branch is designed to address such a limitation by disentangling the texture atlas and geometry translation, producing robust neural texture features capable of handling geometric errors.}

Concretely, the model consists of four main modules: (1) a motion network for producing dense motion flow and translation signals for both appearance and geometry translation, (2) a multi-scale feature warping network for accurate appearance feature alignment in the 2D image plane, (3) a neural texture mapping module for recovering and stabilizing the DensePose annotation by joint training the neural texture atlas translation and DensePose translation, and (4) a blending module for image refinement and super-resolution. Experimental results show that our proposed architecture can benefit from both appearance and geometry information, achieving competitive performance in animating images with the correct pose on three benchmark datasets. Challenging motion patterns, including full body rotation, self-occlusion and subtle hand motions, can be transferred well. Our contributions can be summarized as follows: 

\begin{itemize}
\item  We propose a unified architecture that combines feature warping and neural texture mapping to animate human body images with high-quality pose transfer.
\item  We propose to implicitly disentangle neural texture atlas translation and DensePose translation for better recovery and stabilization of the DensePose geometry under a differentiable neural texture mapping framework.
\item Results on video datasets with half and full human body demonstrate our model can be applied for challenging motion transfer cases, \eg, turning around, self-occlusion, and front-to-back view pose transfer.
\end{itemize}

The rest of this paper is organized as follows. Section \ref{sec:relatedwork} provides a brief introduction to related works for human motion transfer. In Section \ref{sec:method}, we detail our proposed model and highlight the functions of the proposed key modules. The experimental results in comparison with ten state-of-the-art models, as well as ablation studies, are presented in Section \ref{sec:experiments}. Section \ref{sec:discusion} provides a comprehensive discussion and analysis of the limitations of our proposed model. Finally, we conclude the paper and discuss the future outlooks in Section \ref{sec:conclusion}.

\section{Related Work}
\label{sec:relatedwork}

\subsection{Pose-guided Image/Video Generation}
In recent years, GAN-based models have been proposed for human motion transfer. Some works~\cite{nipsWang0ZYTKC18,iccvChanGZE19,nipsWang0TLCK19} treated motion transfer as an aligned image translation task. By leveraging the video-to-video (\texttt{Vid2Vid}) translation framework, models can learn to map the input keypoints or skeletons sequence to photo-realistic human body images. To better recover the body shape, keypoints or skeletons were mapped into semantic masks and then translated into human images \cite{tcybXiaMLZ23,tcsvtYangYZJZS22}. {Specifically, Wang~\etal~\cite{nipsWang0ZYTKC18} introduced the Vid2Vid framework and explored the translation from motion/pose sequence to videos. After that, a few-shot Vid2Vid model is proposed. It can be trained using a small set of video frames of a source person and generalized to new individuals by parameter fine-tuning. Kappel~\etal~\cite{KappelGEHSCTM21} presented a multi-stage framework for achieving person-specific high-fidelity motion transfer. Two consecutive networks, namely shape and structure networks, are adopted to explicitly predict human body parsing and the internal gradient structure of clothing wrinkles, respectively. Albarracin~\etal~\cite{tipAlbarracinR22} developed a self-supervised motion-transfer 
variational autoencoder (VAE). It disentangles the motion and content information from the video chunks and achieves video reenactment by swapping motion features. Sun~\etal~\cite{SunHWLLGpami23} introduced the Laplacian features of the reconstructed 3D mesh as the intrinsic 3D constraints for achieving person-specific human motion transfer. Two motion networks are separately trained for the source and target domain. A detail enhancement network (DE-Net) is leveraged to align the body shape and further transfer the image from the blending domain to the target.}

However, these works commonly constrain motion transfer to a single person and are limited by generalization capabilities.
 Alternatively, the problem can be treated as an unaligned image translation task, by using conditional input like body landmarks \cite{cvprZhuHSYWB19, tmmWeiXSH21}, articulated skeletons \cite{cvprZhuHSYWB19, cvprZhang0L021,tmmWeiXSH21,cvprBhuniaKCALSK23} or semantic masks \cite{nipsDongLGLZY18, RenF0LLcvpr22}. The focus is on establishing the correspondence between source and driving poses via attention mechanisms \cite{cvprZhuHSYWB19,RenF0LLcvpr22} or correlation matrices \cite{Zhang_2020_CVPR,Zhou_2021_CVPR,TaoG0D23flow}. Specifically, Siarohin \etal \cite{nipsSiarohinLT0S19} proposed the first-order motion model (FOMM) based on self-supervised keypoints and 2D affine regional motion.

This was later extended \cite{cvprSiarohinWRCT21} to better disentangle shape and pose, as more appropriate for articulated motion. Zhu \emph{et~al.} \cite{cvprZhuHSYWB19} proposed to progressively transfer a conditional pose through a sequence of intermediate pose representations by leveraging the given source and driving landmarks.
Ren \emph{et~al.} \cite{tipRenLLL20} used a differentiable global-flow local-attention framework to spatially transform inputs at the feature level. 
After that, Ren~\emph{et~al.}~\cite{RenF0LLcvpr22} proposed a double-attention architecture to extract the source appearance feature and distribute the texture to the target pose. By switching the roles of Query, Key, and Values, two cross-attention modules are trained to learn to extract and distribute the feature according to the spatial distribution in the 2D image plane, respectively. {Tao~\etal~\cite{TaoWGJLD22} presented the motion transformer to explicitly model global motion interactions by leveraging query-based vision transformer architecture as a robust motion estimator. Following the unsupervised setting, it achieved object-agnostic image animation.} Men \emph{et~al.} \cite{cvprMenMJML20} proposed a ADGAN. It aims to decompose and embed the human body attributes into a series of latent style codes and achieve explicit appearance control by injecting style codes into different human body parts.
{Wei~\etal~\cite{aaaiWeiXSH21} developed a coarse-to-fine flow warping network by integrating layout synthesis, cloth warping and image composition into a unified framework. Moreover, a Layout-Constrained Deformable Convolution (LC-DConv) layer is proposed to improve spatial consistency, and a Flow Temporal Consistency (FTC) Loss is designed to enhance temporal consistency. Yang~\etal~\cite{YangLLXGY022} proposed a region-to-whole human motion transfer framework, namely REMOT, to progressively align the source appearance texture with the adapted human body semantic layout.}

Pose-guided image generation methods aim to transfer pose for arbitrary persons but do not perform well for animation without temporal modelling. Moreover, such existing works tend to work well only for images within the same dataset, resulting in limited generalization ability \cite{yoon2021pose}. A major challenge is in finding correspondence between source and driving poses due to self-occlusion and missing body parts. Artefacts from unsupervised methods are more significant due to semantic ambiguity in identifying body parts without supervision. In addition, unsupervised 2D-based methods may get confused by whether a person is facing front or back, as 2D landmarks, regions, or semantic masks may not provide adequate guidance of body pose.

\subsection{Neural Rendering Methods}
3D geometry representations for pose transfer can be more effective for  larger pose variations, 
especially for handling occlusion and de-occlusion \cite{yoon2021pose}. Some recent neural rendering-based methods \cite{eccvNeverovaGK18,huang2021few,eccvSarkarMXGT20,bmvcZablotskaiaSZS19,3dimarkarLGT21,yoon2021pose} made use of DensePose \cite{cvprGulerNK18} and parametric 3D models \cite{cvprLiHL19,iccvLiuPML0G19} to improve synthesis quality, such as disentangling the representation of human geometry and texture \cite{eccvSarkarMXGT20}, inpainting a complete set of UV texture from visible parts \cite{eccvNeverovaGK18}, few-shot learning for texture atlas completion \cite{cvprHuangHXZ21}, 3D-based fine-grain appearance flow \cite{yoon2021pose}. Specifically, Neverova \etal~\cite{eccvNeverovaGK18} presented a two-stream pipeline of pose transfer by combining surface-based image synthesis and texture atlas warping. 
Inspired by \cite{thies2019deferred},
Sarkar \etal~\cite{eccvSarkarMXGT20} proposed a neural re-rendering approach to learn a high-dimensional detail-preserving UV feature map,
while Huang \etal~\cite{cvprHuangHXZ21} proposed a few-shot motion transfer method that aims to learn complete texture atlases. Instead of using UV maps for texture mapping, Yang \etal~\cite{tcsvtYangYZJZS22} proposed an SMPL prior-based differentiable render process to map the feature of normalized neural texture atlas to the SMPL model. Sun \etal~\cite{SunHWLLGpami23} proposed a D2G-Net for person-specific motion transfer. The human body keypoints and 2D projection of 3D SMPL mesh are used as the pose representations and fed into a UV generator to produce the IUV map. To recover details, the learnable neural texture combined with RGB texture atlas is adopted for texture mapping.

Our neural texture translation module in our proposed model is closely related to \cite{eccvSarkarMXGT20} and \cite{cvprHuangHXZ21}. However, instead of explicitly extracting the part-based normalized texture atlas in a separate data preparation stage, we proposed to learn the neural texture representation and reduce the impact of inaccurate DensePose IUV maps by decoupling the texture atlas and DensePose translation. Moreover, due to the one-shot setting, the source image can only provide the appearance of a specific single frame. Based on the limited information of the source person and the visibility of human body parts, our proposed model is tailored to learn texture maps, which are conditioned on each driving DensePose, {rather than attempting the ill-posed task of learning to complete the texture atlas by inpainting the invisible parts of the human body.}

\subsection{Diffusion-based Models.}
Recent advances in diffusion-based models have achieved remarkable performance in high-fidelity image generation. The diffusion-based model has been applied to human image animation. Concretely, Bhunia \etal~\cite{cvprBhuniaKCALSK23} proposed the first diffusion-based approach for pose-guided human image synthesis. The target pose and source image, as the conditional inputs, are fed into the denoising diffusion probabilistic model (DDPM)~\cite{nipsHoJA20}. A cross-attention module is applied to query the source texture embedding and inject the style code of the source person according to the target pose. Karras \etal~\cite{corrdreampose23} proposed DreamPose. The CLIP~\cite{icmlRadfordKHRGASAM21} visual embedding of the source image is extracted and then fed into a pre-trained text-to-image stable diffusion model to generate human images. A short sequence of DensePose IUV maps is also provided as the input of the pose condition. However, similar to other diffusion models, the inference time is quite slow. In addition, due to the lack of temporal modelling, the generated videos also show flickers in the background and human body texture, as well as motion blur artefacts~\cite{corrabs-2311-17117}. 

To generate temporally consistency-preserving videos and achieve controllable character animation, Hu~\etal~\cite{corrabs-2311-17117} proposed Animate-Anyone. Different from the ControlNet~\cite{iccv23zhang2023adding}, a ReferenceNet is designed to align the control features and target image. In addition, inspired by \cite{corranimatediff}, a temporal layer is applied to generate temporally smooth video with continuity of appearance details. Instead of using CLIP to produce visual embedding of the reference image \cite{corrabs-2311-17117}, Xu \etal~\cite{corrmagicanimate} proposed another diffusion-based framework, namely MagicAnimate. An AppearnceEncoder is designed to preserve details of the source image. The target DensePose sequence is fed into the ControlNet~\cite{iccv23zhang2023adding} to encode temporal information.  

Different from the above diffusion-based methods, Ni \etal~\cite{cvprNiSLHM23} proposed a text-driven latent flow diffusion model (LFDM) for image-to-video generation. The DDPM is trained to generate the sequence of dense motion flow fields and occlusion maps guided by a given text embedding produced by BERT~\cite{naaclDevlinCLT19}. The UNet-based generator with feature warping~\cite{nipsSiarohinLT0S19,cvprSiarohinWRCT21} is applied to render video frames.

Compared with the above diffusion-based models, our model can run efficiently on lower-end GPU devices. On the other hand, due to the sampling process and strategies, diffusion-based models can not generate temporally consistent and stable videos.

\begin{figure*}[tb!]
  \centering
  \includegraphics[width=0.93\linewidth]{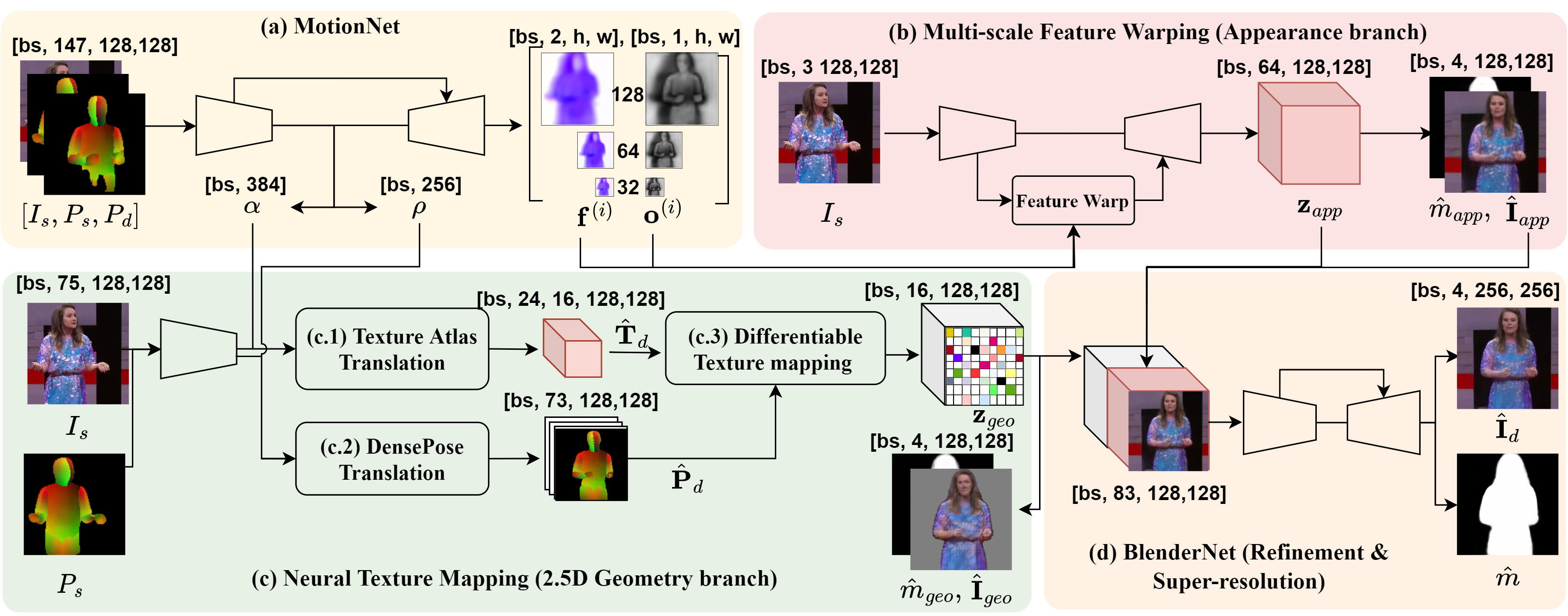}
  \caption{\textbf{Overview of our proposed pipeline}. The MotionNet (a) produces the dense motion flow and translation signals for the appearance and geometry translation branches. We combine multi-scale feature warping (b) and neural texture mapping (c) into a unified framework for motion transfer. The translated appearance and geometry features are integrated by BlenderNet(d) for image refinement.}
  \label{fig1:overview}
\end{figure*}

\section{Method}
\label{sec:method}

Given a static image $\mathbf{I}_s$ and the corresponding DensePose $\mathbf{P}_s$, our main goal is to learn a model $\Phi$ that synthesizes a photorealistic video sequence $\{\hat{\textbf{I}}_d^{(1)},\hat{\textbf{I}}_d^{(2)},\cdots,\hat{\textbf{I}}_d^{(N)}\}$, in a manner that retains the \emph{identity} of the source image $\mathbf{I}_s$ but appropriately drives its \emph{pose} as the dense pose sequence $\{\textbf{P}_d^{(1)},\textbf{P}_d^{(2)},\cdots,\textbf{P}_d^{(N)}\}$ extracted from the target video. 

The overview of our pipeline is shown in Fig.\ \ref{fig1:overview}. In general, there are four major stages involved in our training pipeline. (a) 
A motion network is presented to simultaneously predict the dense flow for 2D appearance feature warping and the translation signals for the 2.5D dense pose branch. (b) A feature warping architecture is applied to capture multi-scale \emph{identity} features w.r.t.\ different scales, while separately (c) a neural texture mapping module is designed to achieve arbitrary neural body image rendering by disentangling the texture and geometry. (d) Finally, we introduce a BlenderNet to further improve the translated image quality by fusing the translated \emph{appearance} and \emph{geometry} features. 

\subsection{Motion Network}\label{sec:motionnet}
Unlike existing methods~\cite{eccvwRenCTFSY20,huang2021few} that directly predict the geometry and appearance from input pose and image, our proposed \texttt{MotionNet} attempts to estimate the residual information of appearance and geometry by accounting for the difference between the target pose $\mathbf{P}_d$ and source pose $\mathbf{P}_s$, \emph{in the context} of the source image $\mathbf{I}_s$. Our \emph{key assumption} is that it should be easier to learn a residual difference between the input and target pose, instead of directly driving the person in a source image. In particular, the proposed \texttt{MotionNet} first estimates a pixel-level dense motion flow for high-quality visual appearance warping in Section~\ref{sec:method_feature}, and it also learns a feature-level translation signals $\mathbf{\alpha}$ and $\mathbf{\rho}$ for the disentangled texture atlas translation and dense pose translation used in Section~\ref{sec:method_neural}. One \emph{notable aspect} is that while $\mathbf{P}_d$ and $\mathbf{P}_s$ establish explicit correspondence between some pixels (though with limited accuracy), these are limited to pixels within the DensePose silhouettes. Our \texttt{MotionNet} learns to exploit this strong guidance to also predict motion flow for pixels \emph{outside} the silhouettes, \eg for hair and clothing, in relation to image features from $\mathbf{I}_s$.

Our \texttt{MotionNet} is built upon FlowNet \cite{cvprIlgMSKDB17}  (Fig.~\ref{fig1:overview} (a)), in which an encoder 
is responsible for extracting feature-level translation signals for the 2.5D geometry branch, and a decoder has an iteratively upsampled refinement framework to predict multi-scale pixel-level dense motion flows and occlusion maps. This can be expressed as:
\begin{equation}
  \left[ \mathbf{\alpha}, \mathbf{\rho}, [\mathbf{f}^{(i)}, \mathbf{o}^{(i)}] \right]=\texttt{MotionNet}(\mathbf{I}_s, \mathbf{P}_s, \mathbf{P}_d),
  \label{eq_motionnet}
\end{equation}
where $\mathbf{\alpha}$ and $\mathbf{\rho}$ respectively provide guidance for neural texture atlas translation and dense pose translation, while $\mathbf{f}^{(i)}$ and $\mathbf{o}^{(i)}$ respectively denote the predicted $i$-th level dense motion flow and occlusion map. 

\subsection{Feature Warping Network}\label{sec:method_feature}
After estimating the multiscale dense motion flow $\mathbf{f}$ and occlusion map $\mathbf{o}$ using the \texttt{MotionNet}, we would like to predict the more refined 2D transferred appearance by aligning the source image content based on these initial transfer signals. In particular, we implement this function using a multi-level feature framework shown in Fig.~\ref{fig1:overview}(b), which can capture different scales of motion on different feature maps. More specifically, given a source image $\mathbf{I}_s$, we first embed it in multi-scale features through:
\begin{equation}
  \left[\textbf{z}_s^{(1)}, \textbf{z}_s^{(2)},\cdots, \textbf{z}_s^{(L)} \right]=\mathcal{E}_s(\mathbf{I}_s),
  \label{eq_es}
  \end{equation}
where $\textbf{z}^{(i)}_s$ denotes the $i$-th level source feature embedded by the encoder $\mathcal{E}_s$.
In conjunction with the multi-scale motion flow $\mathbf{f}$ obtained from \texttt{MotionNet}, 
our main goal is then to \emph{warp the source feature through the residual motion flow}. Furthermore, we also propose to use the additional occlusion maps $\mathbf{o}$ as filters to propagate the source information to the transferred feature, \ie, when the occlusion is 0, source-aligned features will be directly passed; otherwise the previously warped features will be fed in. Formally, the operation is defined as:
\begin{equation}
\hat{\mathbf{z}}_s^{(i+1)}=\texttt{Warp}(\mathbf{z}_s^{(i)}, \mathbf{f}^{(i)})*\mathbf{o}^{(i)}+\hat{\mathbf{z}}_s^{(i)}*(1-\mathbf{o}^{(i)}),
\label{eq_fw}
\end{equation}
where $\texttt{Warp}$ is the flow warping function, which in practice can be implemented by a bilinear sampling operation, while
the occlusion map $\mathbf{o}^{(i)}$ determines how the warped source feature should be combined with the current aligned feature $\hat{\mathbf{z}}_s^{(i)}$.
The final fused appearance feature $\mathbf{z}_{app}$ is used to generate a coarse image.
In addition, this stage also predicts a human segmentation mask that enables extra supervision during training, with a prediction head in the generator.
The final output of the 2D feature warping branch is:
\begin{equation}
  \left[ \hat{\mathbf{I}}_{app}, \hat{m}_{app} \right]=\mathcal{G}_{app}(\mathbf{z}_{app}),
  \label{eq_gen2d}
\end{equation}
\noindent where $\hat{\mathbf{I}}_{app}$ is the generated image and $\hat{m}_{app}$ denotes the predicted binary mask.

\subsection{Neural Texture Mapping}\label{sec:method_neural}
At the core of our pipeline is a neural texture mapping module, which learns to implicitly disentangle the texture mapping with neural texture atlas (\emph{Appearance}) and DensePose IUV map (\emph{Geometry}) translation. Note that we bypass the existing neural rendering-based methods \cite{thies2019deferred,cvprRajTHVSL21,eccvSarkarMXGT20} with the following three unique considerations: 1) The existing neural texture mapping methods decompose the correspondence of appearance and geometry using the DensePose, which is based on the SMPL model~\cite{togLoperM0PB15}, that strictly extracts the pixels within the SMPL silhouette, ignoring other parts, such as the hairs and background. 2) Errors in the extracted DensePose IUV map, including occasional large missing gaps, have a significant negative effect on the corresponding extracted textures. 3) Direct extraction of texture atlases or part-based normalized textures based on DensePose correspondence may not only be time-consuming (unless carefully GPU-optimized), but the quality of the extracted texture atlases may also be poor due to the limited resolution accuracy of DensePose.

\paragraph{\textbf{Texture Atlas and DensePose Decoupling}} We present a two-stream neural texture mapping framework shown in Fig.~\ref{fig1:overview}(c), in which we attempt to tame our model to implicitly decouple the translation of the neural texture atlas and DensePose IUV map. In particular, given a source image $\mathbf{I}_s$, as well as the corresponding extracted DensePose map $\textbf{P}_s$, and translation signals $\mathbf{\alpha}$, $\mathbf{\rho}$ (extracted from \texttt{MotionNet}), the texture atlas and geometry translation are respectively formulated as:
\begin{equation}
\hat{\mathbf{T}}_d =\mathcal{T}_{app}(\mathcal{E}_{app}(\mathbf{I}_s, \mathbf{P}_s), \mathbf{\alpha}),\: \hat{\mathbf{P}}_d=\mathcal{T}_{geo}(\mathcal{E}_{geo}(\mathbf{I}_s, \mathbf{P}_s), \mathbf{\rho}),
\label{eq_ntt}
\end{equation}
where $\mathcal{E}_{app}$ and $\mathcal{E}_{geo}$ are the appearance encoder and geometry encoder, respectively. They have some initial shared weights, but with two separate convolutional layers. $\mathcal{T}_{app}$ and $\mathcal{T}_{geo}$ refer to translation networks that output the neural texture atlas $\hat{\mathbf{T}}_d \in \mathbb{R}^{24\times c\times h \times w}$ and the translated DensePose IUV map $\hat{\mathbf{P}}_d$, respectively. In detail, the translated DensePose $\hat{\mathbf{P}}_d$ consists of three components, \ie $[\hat{\mathbf{U}}, \hat{\mathbf{V}}, \hat{\mathbf{S}}]$, where $\hat{\mathbf{U}}\in\mathbb{R}^{24 \times h \times w}$ and $\hat{\mathbf{V}}\in\mathbb{R}^{24 \times h \times w}$ are the predicted UV coordinates for 24 non-overlapping body parts defined in DensePose, with spatial size $h\times w$, while $\hat{\mathbf{S}} \in \mathbb{R}^{25 \times h \times w}$ represents the semantic score map for the different parts. In practice, a $\texttt{softmax}$ function is applied to estimate the probabilities of each pixel belonging to different body parts. The $0$-th channel with background probability in $\hat{S}$ is ignored in the differentiable texture mapping process.

\paragraph{\textbf{Differentiable Texture Mapping}} Once we predict the translated neural texture atlas feature and the soft mask for each body part, we propose to use a \emph{differentiable texture mapping} module to produce the target neural texture feature, which is subsequently rendered to a coarse output. First, we compute:
\begin{equation}
\hat{R}^{(k)}_{i, j}=\hat{\mathbf{T}}_d^{(k)}\left(\hat{\mathbf{U}}^{(k)}_{i,j}, \hat{\mathbf{V}}^{(k)}_{i,j}\right),
\label{eq_ntm}
\end{equation}
where $i$ and $j$ index to the predicted UV coordinates for the $k$-th body part. Once the UV coordinates are obtained, the corresponding neural texture feature is assigned to the $[i,j]$ location. In practice, the above process can be efficiently implemented using a differentiable bilinear sampling operation, in which the $k$-th component of UV coordinates will be regarded as a sampling grid, and the $k$-th neural texture atlas feature is the input for sampling. After that, neural texture feature maps for different body parts $\hat{R}^{(k)}$ 
are weighted summed according to the probability score map $\hat{S}$: 
\begin{equation}
  \mathbf{z}_{geo}=\sum_{k=1}^{24} \hat{S}^{(k)}\hat{R}^{(k)},
  \label{eq_wsall}
\end{equation}
\noindent where $\mathbf{z}_{geo}\in\mathbb{R}^{c\times h \times w}$ is the unified neural texture feature. 

\paragraph{\textbf{Coarse Transferred Motion}} Although we implicitly decouple the neural texture mapping and DensePose IUV map in a \emph{differentiable} module, we do \emph{not} have the ground truth for both parts. Hence, an important issue to address is: how can we leverage the 2D image to guide the learning of this neural texture mapping?
We tackle this by first rendering a coarse image using the neural texture feature map with a single convolutional layer. Following \cite{cvprRajTHVSL21}, we also introduce a foreground human body mask as extra supervision to guide the training, to improve UV completion and extraction of neural texture features. In our implementation, we adopted two separate prediction heads to generate the coarse image and predict the foreground mask:
\begin{equation}
  \left[ \hat{\mathbf{I}}_{geo}, \hat{m}_{geo} \right]=\mathcal{G}({\mathbf{z}}_{geo}),
  \label{eq_geoout}
\end{equation}
where $\hat{\mathbf{I}}_{geo}$ is the rendered coarse image and $\hat{m}_{geo}$ the predicted binary mask.

\subsection{Blending Network}\label{sec:method_blender}
Finally, a shallow blending network is applied to further improve the visual quality of transferred results by fusing features from the \emph{2D appearance} flow warping and \emph{2.5D geometry} neural texture mapping modules. Following ANR~\cite{cvprRajTHVSL21}, we simultaneously predict the final translated RGB image $\hat{\textbf{I}}_{d}$ and the corresponding binary mask $\hat{m}$ as:
\begin{equation}
  \left[ \hat{\textbf{I}}_d, \hat{m} \right]=\texttt{BlenderNet}(\textbf{z}_{geo}, \textbf{z}_{app}, \hat{\textbf{I}}_{app}).
  \label{eq_blender}
\end{equation}

\subsection{Loss functions}
In our implementation, multiple loss functions are used in training our whole pipeline, which consists of four different branches.

\paragraph{\textbf{Perceptual Correctness Loss and Regularization Terms}} To stabilize the dense motion estimation module in the 2D branch, we first used the perceptual correctness loss \cite{tipRenLLL20} and an affine transformation regularization term \cite{tipRenLLL20} to penalize the error dense flow field. In addition, a multi-scale consistency loss is introduced to preserve the feature warping consistency among different scales. We have also introduced a total-variation (TV) regularization term to smooth the estimated dense motion flow in each level of output of the \texttt{MotionNet}.
\begin{equation}
  \begin{aligned}
   \mathcal{L}_{\mathrm{motion}}=&\lambda_{cor}\mathcal{L}_{correctness} + \lambda_{reg}\mathcal{L}_{regularization} \\
   &+ \lambda_{tv}\mathcal{L}_{TV} + \lambda_{con}\mathcal{L}_{consistency}.
  \end{aligned}
  \label{eq_motion}
  \end{equation}
For TV loss and consistency loss, we calculate the TV loss in each scale of dense motion flow to regularize the model for producing smooth flow. For multi-scale consistency loss, we use the flow with the lowest resolution (i.e., 32x32 in our implementation) as the soft label to calculate the consistency loss with flows with higher resolutions (i.e. 64x64, and 128x128 in our implementation). In our experiments, the weight hyperparameters $\lambda_{cor}, \lambda_{reg}, \lambda_{tv}$ and $\lambda_{con}$ are set to 5.0, 0.01, 1.0 and 5.0, respectively.

\paragraph{\textbf{DensePose Estimation Loss and Masked L1 Loss}} For the neural texture mapping module in the 2.5D branch, we adopted the cross-entropy (CE) loss for the semantic mask prediction of the DensePose IUV map and a masked L1 loss for UV coordinates regression \cite{cvprHuangHXZ21}. For the neural rendered RGB image, since the differentiable neural texture mapping is based on the correspondence between the neural texture atlas and UV coordinates, pixels outside the predicted DensePose silhouettes should not be considered for calculating L1 reconstruction loss. We thus applied masked L1 loss to filter out the regions outside the human regions. 
\begin{equation}
\begin{aligned}
  \mathcal{L}_{iuv}&=\lambda_{uv}\sum_{k=1}^{24}(\left\|\hat{\mathbf{S}}^{(k)} \odot (\mathbf{U}^{(k)} - \hat{\mathbf{U}}^{(k)}) \right\|_{1} \\ &+\left\|\hat{\mathbf{S}}^{(k)} \odot (\mathbf{V}^{(k)} - \hat{\mathbf{V}}^{(k)}) \right\|_{1})+\lambda_{ce}\mathcal{L}_{CE}({\mathbf{S}}^{(k)}, \hat{\mathbf{S}}^{(k)}),
\end{aligned}
    \label{eq_iuvloss}
\end{equation}
where $\hat{\mathbf{U}}$ and $\hat{\mathbf{V}}$ denote the predicted UV components of the 24 human body parts. $\hat{\mathbf{S}}$ represents the probability score map. $[\mathbf{U}, \mathbf{V}, \mathbf{S}]$ are the pseudo ground-truths provided by the DensePose IUV map. In our implementation, $\lambda_{uv}$ and $\lambda_{ce}$ are set to 5.0, and 1.0, respectively.

\paragraph{\textbf{Image Pyramid Perceptual Loss and Reconstruction loss}} Following \cite{cvprSiarohinWRCT21,nipsSiarohinLT0S19,cvprHuangHXZ21,eccvSarkarMXGT20}, an image pyramid perceptual loss and L1 reconstruction loss are introduced to get better image quality. In our implementation, the perceptual loss and reconstruction loss are adopted for the generated RGB image of the 2D branch and \texttt{BlenderNet}.
\begin{equation}
\begin{aligned}
  \mathcal{L}_{rec}&=\lambda_{p}\mathcal{L}_{\mathrm{p}}(\mathbf{I}_d, \hat{\mathbf{I}}_{\varrho })+\lambda_{1}\mathcal{L}_{\mathrm{1}} \\ 
  &=\lambda_{p}\sum_{l=1}^{L}\sum_{i=1}^{N}\left\|{\Phi}^{i}({\Omega}^{l}(\mathbf{I}_d))-{\Phi}^{i}({\Omega}^{l}(\hat{\mathbf{I}}_{\varrho }))\right\|_{2}+\lambda_{1}\|\mathbf{I}-\hat{\mathbf{I}}_{\varrho }\|_{1},
\end{aligned}
    \label{eq_ppl}
\end{equation}
where $\varrho =\{app, geo, d\}$ denotes a set of images generated by different modules, $\mathbf{I}_d$ is the ground-truth image. $\Omega^{l}$ is the $l$-th scale of image resized by using a down-sampling operation. $\Phi^{i}$ is the $i$-th layer of VGG-19 pretrained network. 
In our experiments, $\lambda_{p}$ and $\lambda_{1}$ are set to 10.0, and 1.0, respectively.

\paragraph{\textbf{Binary Cross-Entropy (BCE) Loss}} Inspired by \cite{cvprRajTHVSL21}, a human foreground mask prediction head is adopted as extra guidance for the 2D, 2.5D and blending modules. Concretely, the BCE loss in the 2D branch can not only be used to penalize the error in the estimated motion flow but also encourage the model to recover the appearance in the human regions. For the neural texture mapping branch, the human foreground mask can provide extra supervision to guide the generation of content outside the DensePose silhouettes but inside the boundary of the human foreground mask. 
\begin{equation}
  \mathcal{L}_{mask}({m},\hat{{m}}) = \texttt{BCE} ({m},\hat{m}_{\varrho }),
  \label{eq_mask}
\end{equation}
where ${m}$ is the pseudo ground-truth mask, and $\hat{m}_{\varrho }$ the human mask predicted by the different modules.

\paragraph{\textbf{Adversarial Loss}} Finally, we adopted the Least-square GAN (LSGAN) loss \cite{iccvMaoLXLWS17} to further improve the quality of generated images.
\begin{equation}
\begin{aligned}
  \mathcal{L}_{{adv}}(D, G)&=\frac{1}{2} \mathbb{E}[(D({I_d})-1)^2]+\frac{1}{2} \mathbb{E}[D(G(I_s, P_s, P_d))^2] \\ &+\frac{1}{2} \mathbb{E}[(D(G(I_s, P_s, P_d))-1)^2],
\end{aligned}
  \label{eq_lsgan}
\end{equation}
where $\mathcal{D}$ denotes the Discriminator. In our implementation, we adopted a simple Discriminator with ResNet blocks.

\section{Experiments}
\label{sec:experiments}

\noindent\textbf{Datasets.} We trained and evaluated our proposed model on three video benchmark datasets for human motion transfer, including \textbf{TED-Talks} \cite{cvprSiarohinWRCT21}, \textbf{TaiChiHD} \cite{nipsSiarohinLT0S19}, and \textbf{iPER} \cite{iccvLiuPML0G19}. \textbf{TED-Talks} contains 1,123 video clips for training and 131 clips for testing, which are segmented from TED Talks Youtube videos. In each video, the upper part of the speaker is cropped from the video according to the annotated bounding boxes. \textbf{TaiChiHD} consists of 2,927 video clips for training and 253 clips for testing, which capture full-bodied tai chi actions. \textbf{iPER} dataset consists of 206 video sequences. Among them, there are 164 and 42 videos for training and testing, respectively. The video sequences are relatively longer than TED-Talks and TaiChiHD, and include A-poses and random actions. In our experiments, for each dataset, we cropped and resized the video frames to 256$\times$256. To obtain pseudo data annotation, we used the latest released DensePose model \cite{cvprGulerNK18} to get the IUV map for video frames in the above three datasets. The model uses ResNet101 \cite{cvprHeZRS16} as the backbone and a DeepLabV3 \cite{corrChenPSA17} prediction head. Moreover, we adopted a video human matting method \cite{wacvLinYSS22} to obtain the human foreground masks as pseudo-ground truth for training.

\noindent\textbf{Evaluation Settings.} We evaluated the performance on four quantitative metrics, including \textit{L1} error, \textit{Fréchet Inception Distance} (FID)~\cite{fidHeuselRUNH17}, \textit{Average Euclidean Distance} (AED), \textit{Average Keypoint Distance} (AKD), and \textit{Missing Keypoint Rate} (MKR)~\cite{cvprSiarohinWRCT21}. Specifically, the L1 error measures the difference between the generated frames and ground-truth videos. Since we typically do \emph{not} have the ground-truth animation videos, the L1 error can only be applied to evaluate the self-reconstruction of test videos. AED evaluates the preservation of identity before and after motion transfer. We used the person re-identification model \cite{corrHermansBL17} for extracting identity features, and calculated the mean L2 norm of the identity features between the generated and ground-truth frame pairs. AKD and MKR indicate how well the pose is transferred in the reconstructed videos. The keypoints of the ground-truth and generated videos are estimated by using a publicly available pose estimation model \cite{cvprCaoSWS17}. The AKD calculates the average distance between the corresponding human body landmarks, and the MKR is the missing rate of the landmarks by checking the presence of keypoints for a pair of ground-truth and generated frames. 

\begin{table*}[htbp]
  \centering
  \caption{Quantitative results for self-reconstruction in comparison with state-of-the-art models.}
  \scalebox{0.86}{
    \begin{tabular}{l|ccccc|ccccc|ccccc}
      \hline
            & \multicolumn{5}{c|}{TED-Talks}        & \multicolumn{5}{c|}{TaiChiHD}         & \multicolumn{5}{c}{iPER} \\
  \cline{2-16}          & L1↓   & FID↓  & AED↓  & MKR↓  & AKD↓  & L1↓   & FID↓  & AED↓  & MKR↓  & AKD↓  & L1↓   & FID↓  & AED↓  & MKR↓  & AKD↓ \\
      \hline
      FOMM & 0.0303 & 23.6767 & 0.1424 & 0.0102 & 3.8112 & 0.0606 & \cellcolor[rgb]{ .851,  .851,  .851}24.7575 & 0.1667 & 0.0299 & 6.5769 & 0.0226 & 28.3101 & \cellcolor[rgb]{ .851,  .851,  .851}0.0689 & 0.0102 & 2.1754 \\
      GFLA & 0.0597 & 33.3915 & 0.2353 & 0.0243 & 7.3364 & 0.1002 & 31.6874 & 0.2246 & 0.0174 & 5.9699 & 0.0277 & 39.4560 & 0.1016 & 0.0066 & 1.8302 \\
      MRAA & \cellcolor[rgb]{ .749,  .749,  .749} 0.0264 & \cellcolor[rgb]{ .651,  .651,  .651}\textbf{17.7006} & \cellcolor[rgb]{ .749,  .749,  .749}0.1166 & 0.0086 & 2.6573 & \cellcolor[rgb]{ .851,  .851,  .851}0.0462 & 28.2660 & \cellcolor[rgb]{ .851,  .851,  .851}0.1507 & 0.0263 & 5.2372 & \cellcolor[rgb]{ .651,  .651,  .651}\textbf{0.0166} & \cellcolor[rgb]{ .749,  .749,  .749}20.8602 & \cellcolor[rgb]{ .651,  .651,  .651}\textbf{0.0567} & 0.0068 & 1.5992 \\
      TPSMotion & \cellcolor[rgb]{ .851,  .851,  .851}0.0272 & \cellcolor[rgb]{ .851,  .851,  .851}18.3020 & 0.1245 & \cellcolor[rgb]{ .749,  .749,  .749}0.0073 & \cellcolor[rgb]{ .851,  .851,  .851}2.3684 & \cellcolor[rgb]{ .749,  .749,  .749}0.0456 & \cellcolor[rgb]{ .651,  .651,  .651}\textbf{23.2794} & \cellcolor[rgb]{ .749,  .749,  .749}0.1513 & 0.0182 & \cellcolor[rgb]{ .851,  .851,  .851}4.4937 & 0.0197 & 27.0252 & 0.0874 & 0.0066 & 1.5291 \\
      IAPM & 0.0275 & 26.6549 & \cellcolor[rgb]{ .851,  .851,  .851}0.1220 & 0.0099 & 3.3623 & 0.0523 & 25.2084 & 0.1523 & 0.0240 & 5.6255 & \cellcolor[rgb]{ .749,  .749,  .749}0.0188 & 25.5329 & 0.0733 & 0.0079 & 1.8522 \\
      DAM & \cellcolor[rgb]{ .651,  .651,  .651}\textbf{0.0260} & 18.4181 & \cellcolor[rgb]{ .651,  .651,  .651}\textbf{0.1138} & \cellcolor[rgb]{ .851,  .851,  .851}0.0080 & 2.6727 & \cellcolor[rgb]{ .651,  .651,  .651}\textbf{0.0452} & \cellcolor[rgb]{ .749,  .749,  .749}23.3396 & \cellcolor[rgb]{ .651,  .651,  .651}\textbf{0.1481} & 0.0219 & 5.1399 & \cellcolor[rgb]{ .851,  .851,  .851}0.0189 & \cellcolor[rgb]{ .851,  .851,  .851}24.9596 & \cellcolor[rgb]{ .749,  .749,  .749}0.0622 & 0.0065 & 1.8065 \\
      LIA & 0.0328 & \cellcolor[rgb]{ .749,  .749,  .749}18.0319 & 0.1297 & 0.0089 & 3.0617 & 0.0698 & 30.6238 & 0.1978 & 0.0217 & 6.3817 & 0.0342 & 47.3365 & 0.1263 & 0.0160 & 2.9015 \\
      NTED & 0.0685 & 59.0862 & 0.1327 & 0.0112 & 3.1567 & 0.1234 & 66.6038 & 0.3835 & \cellcolor[rgb]{ .651,  .651,  .651}\textbf{0.0134} & \cellcolor[rgb]{ .749,  .749,  .749}3.3641 & 0.0421 & 45.6141 & 0.1326 & \cellcolor[rgb]{ .651,  .651,  .651}\textbf{0.0045} & \cellcolor[rgb]{ .651,  .651,  .651}\textbf{0.8663} \\
      PIDM & 0.2189 & 64.9514 & 0.2345 & 0.0350 & 3.8160 & 0.1442 & 32.1372 & 0.2487 & 0.0480 & 6.4013 & 0.0622 & 55.0202 & 0.1552 & 0.0065 & 1.1789 \\
      MagicAnimate & 0.0971 & 28.1062 & 0.2420 & 0.0133 & 3.7633 & 0.0907 & 31.2045 & 0.2304 & 0.0195 & 6.4062 & 0.0491 & 30.0272 & 0.1235 & \cellcolor[rgb]{ .749,  .749,  .749}0.0054 & 4.4723 \\
      \hline
      Ours(128$\rightarrow$256) & 0.0316 & 26.3862 & 0.1478 & \cellcolor[rgb]{ .651,  .651,  .651}\textbf{0.0071} & \cellcolor[rgb]{ .749,  .749,  .749}2.1255 & 0.0593 & 30.3241 & 0.1764 & \cellcolor[rgb]{ .851,  .851,  .851}0.0137 & 3.6630 & 0.0206 & 25.9149 & 0.0828 & 0.0056 & \cellcolor[rgb]{ .851,  .851,  .851}1.1591 \\
      Ours(256$\rightarrow$256) & 0.0317 & 20.1502 & 0.1423 & 0.0085 & \cellcolor[rgb]{ .651,  .651,  .651}\textbf{1.9866} & 0.0579 & 25.0364 & 0.1774 & \cellcolor[rgb]{ .749,  .749,  .749}0.0141 & \cellcolor[rgb]{ .651,  .651,  .651}\textbf{3.0562} & 0.0201 & \cellcolor[rgb]{ .651,  .651,  .651}\textbf{20.4217} & 0.0860 & \cellcolor[rgb]{ .851,  .851,  .851}0.0055 & \cellcolor[rgb]{ .749,  .749,  .749}0.9184 \\
      \hline
      \end{tabular}%
}
  \label{tab:selfrec-sota}%
\end{table*}%

\noindent\textbf{Experimental Setup and Implementation Details.} To stabilize the training process, our model is trained with a multi-stage setting. Specifically, the neural texture mapping branch (geometry branch) and the encoder of \texttt{MotionNet} are first trained. Then, the \texttt{MotionNet} and multi-scale feature warping branch are trained in the second stage after freezing the other parts of the model. Finally, we trained the full model end-to-end. During training, we optimized the model by using the Adam optimizer \cite{corrKingmaB14} 
with the learning rate as 2e-4. For each dataset, we trained the model with the same hyper-parameter settings. More specifically, the lengths of latent feature vectors of the texture atlas translation signal and the DensePose translation signal are set to 384 and 256, respectively.
We fixed the number of training iterations to 600,000. In the first training stage, we trained the neural texture mapping branch (geometry branch) with 100,000 iterations. Then we trained the multi-scale feature warping branch in the second stage with another 100,000 iterations. Then we trained the full model with 200,000 iterations end-to-end and fine-tuned the BlenderNet with a super-resolution block with 200,000 iterations. 
Our models for ablation analysis were trained by setting the batch size to 8. The input and output resolutions were set to 128x128 and 256x256, respectively. All of the models for comparison with state-of-the-art and ablation studies were trained and tested on a workstation with an NVIDIA GeForce RTX 3090Ti (24G) GPU. We refer the readers to the \href{https://github.com/AndrewChiyz/ORFPNT_supplementary/blob/main/suppledoc.pdf}{supplementary document} for more details on model structure.

\subsection{Main results} 
We first compared our proposed model with state-of-the-art (SOTA) methods for human image animation, including FOMM \cite{nipsSiarohinLT0S19}, MRAA \cite{cvprSiarohinWRCT21}, GFLA~\cite{tipRenLLL20}, DAM~\cite{TaoWXGJLD22}, IAPM~\cite{cvprShalevW22}, TPSMotion~\cite{cvprZhaoZ22}, LIA~\cite{iclrWangYBD22}, NTED~\cite{RenF0LLcvpr22}, PIDM~\cite{cvprBhuniaKCALSK23} and MagicAnimate~\cite{corrmagicanimate}. For a fair comparison, we retrained these SOTA models on the three benchmark datasets at 256x256 resolution. 
All the models, except MagicAnimate~\cite{corrmagicanimate}, were trained under the default hyper-parameter settings of the training schedule as provided in the released code. For MagicAnimate~\cite{corrmagicanimate}, we use the released model to generate videos directly without training and fine-tuning on the above datasets. Following \cite{cvprSiarohinWRCT21}, we evaluated the performance and video quality both on self-reconstruction and cross-video animation.

\begin{table*}[tb!]
  \centering
  \caption{Quantitative comparison and human evaluation for cross-video animation in comparison with SOTA models.}
  \scalebox{0.95}{
    \begin{tabular}{l|ccc|ccc|ccc|cccc}
      \hline
            & \multicolumn{3}{c|}{TED-Talks} & \multicolumn{3}{c|}{TaiChiHD} & \multicolumn{3}{c|}{iPER} & \multicolumn{4}{c}{Human Evaluation} \\
  \cline{2-14}          & AED↓  & MKR↓  & AKD↓  & AED↓  & MKR↓  & AKD↓  & AED↓  & MKR↓  & AKD↓  & \textit{Cont.}↑ & \textit{App.}↑ & \textit{Geom.}↑ & \textit{Overall}↑ \\
      \hline
      FOMM  & \cellcolor[rgb]{ .651,  .651,  .651}\textbf{0.2484} & 0.0496 & 24.1217 & \cellcolor[rgb]{ .851,  .851,  .851}0.2337 & 0.0776 & 19.5635 & \cellcolor[rgb]{ .651,  .651,  .651}\textbf{0.1518} & 0.0193 & 11.2709 & 2.364 & 2.636 & 1.981 & 2.142 \\
      GFLA  & \cellcolor[rgb]{ .851,  .851,  .851}0.3229 & 0.0320 & 11.3251 & 0.2984 & \cellcolor[rgb]{ .749,  .749,  .749}0.0166 & 6.8594 & 0.2097 & 0.0062 & 2.3531 & 2.778 & 2.144 & 2.675 & 2.467 \\
      MRAA  & 0.3391 & \cellcolor[rgb]{ .749,  .749,  .749}0.0159 & 7.2375 & 0.2988 & 0.0542 & 12.0688 & 0.2076 & 0.0207 & 5.7536 & 2.294 & 2.144 & 2.161 & 2.050 \\
      TPSMotion & \cellcolor[rgb]{ .749,  .749,  .749}0.2580 & 0.0589 & 23.9190 & \cellcolor[rgb]{ .749,  .749,  .749}0.2285 & 0.0589 & 18.8978 & \cellcolor[rgb]{ .749,  .749,  .749}0.1674 & 0.0126 & 10.9666 & 2.606 & \cellcolor[rgb]{ .851,  .851,  .851}2.761 & 2.172 & 2.342 \\
      IAPM  & 0.3380 & 0.0189 & 6.8729 & 0.2916 & 0.0322 & 8.9685 & 0.2072 & 0.0176 & 4.2116 & 2.861 & 2.292 & 2.692 & 2.519 \\
      DAM   & 0.3563 & \cellcolor[rgb]{ .651,  .651,  .651}\textbf{0.0143} & 5.7018 & 0.3234 & 0.0338 & 8.2810 & 0.2667 & 0.0227 & 4.3494 & \cellcolor[rgb]{ .851,  .851,  .851}2.867 & 2.361 & 2.756 & 2.531 \\
      LIA   & 0.3418 & 0.0195 & 6.4688 & \cellcolor[rgb]{ .651,  .651,  .651}\textbf{0.2246} & 0.0577 & 18.8209 & \cellcolor[rgb]{ .851,  .851,  .851}0.1841 & 0.0265 & 12.0901 & 2.636 & 2.311 & 2.317 & 2.264 \\
      NTED  & 0.4187 & \cellcolor[rgb]{ .851,  .851,  .851}0.0174 & \cellcolor[rgb]{ .651,  .651,  .651}\textbf{3.1659} & 0.4486 & \cellcolor[rgb]{ .651,  .651,  .651}\textbf{0.0144} & \cellcolor[rgb]{ .651,  .651,  .651}\textbf{2.8236} & 0.2604 & \cellcolor[rgb]{ .651,  .651,  .651}\textbf{0.0045} & \cellcolor[rgb]{ .651,  .651,  .651}\textbf{0.9697} & 2.669 & 2.394 & \cellcolor[rgb]{ .851,  .851,  .851}2.950 & \cellcolor[rgb]{ .851,  .851,  .851}2.581 \\
      PIDM  & 0.3849 & 0.0491 & \cellcolor[rgb]{ .851,  .851,  .851}4.8909 & 0.3347 & 0.0615 & 7.8160 & 0.2645 & 0.0073 & \cellcolor[rgb]{ .851,  .851,  .851}1.3276 & 1.567 & 1.697 & 2.108 & 1.589 \\
      MagicAnimate & 0.3926 & 0.0204 & 5.7672 & 0.3208 & 0.0188 & 7.8961 & 0.2128 & \cellcolor[rgb]{ .749,  .749,  .749}0.0049 & 4.3686 & 2.183 & 1.844 & 2.153 & 1.881 \\
      \hline
      Ours(128→256) & 0.3515 & 0.0186 & 3.6755 & 0.3097 & 0.0171 & \cellcolor[rgb]{ .851,  .851,  .851}3.9812 & 0.2128 & \cellcolor[rgb]{ .851,  .851,  .851}0.0053 & 1.4138 & \cellcolor[rgb]{ .651,  .651,  .651}\textbf{3.617} & \cellcolor[rgb]{ .749,  .749,  .749}2.922 & \cellcolor[rgb]{ .651,  .651,  .651}\textbf{3.719} & \cellcolor[rgb]{ .651,  .651,  .651}\textbf{3.511} \\
      Ours(256→256) & 0.3514 & 0.0193 & \cellcolor[rgb]{ .749,  .749,  .749}3.5150 & 0.3219 & \cellcolor[rgb]{ .851,  .851,  .851}0.0165 & \cellcolor[rgb]{ .749,  .749,  .749}3.0181 & 0.2173 & \cellcolor[rgb]{ .851,  .851,  .851}0.0053 & \cellcolor[rgb]{ .749,  .749,  .749}1.1343 & \cellcolor[rgb]{ .749,  .749,  .749}3.397 & \cellcolor[rgb]{ .651,  .651,  .651}\textbf{2.972} & \cellcolor[rgb]{ .749,  .749,  .749}3.603 & \cellcolor[rgb]{ .749,  .749,  .749}3.431 \\
      \hline
      \end{tabular}%
  }
  \label{tab:animation-sota}%
\end{table*}%

\textbf{1) Video Self-Reconstruction:} We evaluated the quality of generated videos for self-reconstruction, with source and driving frames taken from the same video (and hence the same identity). The first frame of a test video was selected as the source frame, while the other frames of the same video were utilized as the driving frames. Under such a setting, we have ground truth for synthesized frames. 

We compare our approach to the SOTA methods in Table~\ref{tab:selfrec-sota}. The three best results are highlighted using a dark-to-light gradient colormap, where dark represents the best and light is the third best. It shows that our proposed model achieved mixed but generally superior performance on pose transfer, with AKD and MKR values lower than the other methods. However, our model shows slightly higher values on L1 and AED, which relate to image quality and identity preservation, respectively. We believe there are two main reasons for that: 1) Resolution for training. Due to the structure of the main components of our model and having limited GPU memory, we used 128$\times$128 as input, but 256$\times$256 as output under the training scheme. It meant that the model needed to predict the higher-resolution appearance of the human body regions from lower-resolution input. 
2) Unstable DensePose annotation. Since the IUV maps are predicted frame by frame, the input IUV map of the model can be unstable across the video sequence. Consequently, the generated frames will also slightly deviate from the ground truth. We believe the performance of image quality in terms of L1 and AED can be further improved by solving the above issues.
For the diffusion-based model PIDM~\cite{cvprBhuniaKCALSK23}, the model cannot sample temporally consistent and stable videos. Therefore, the performance is lower than ours. In practice, it may take more time with more samples to train the diffusion-based models for high-quality video generation.

\begin{figure*}
  \centering
  \includegraphics[width=1.0\linewidth]{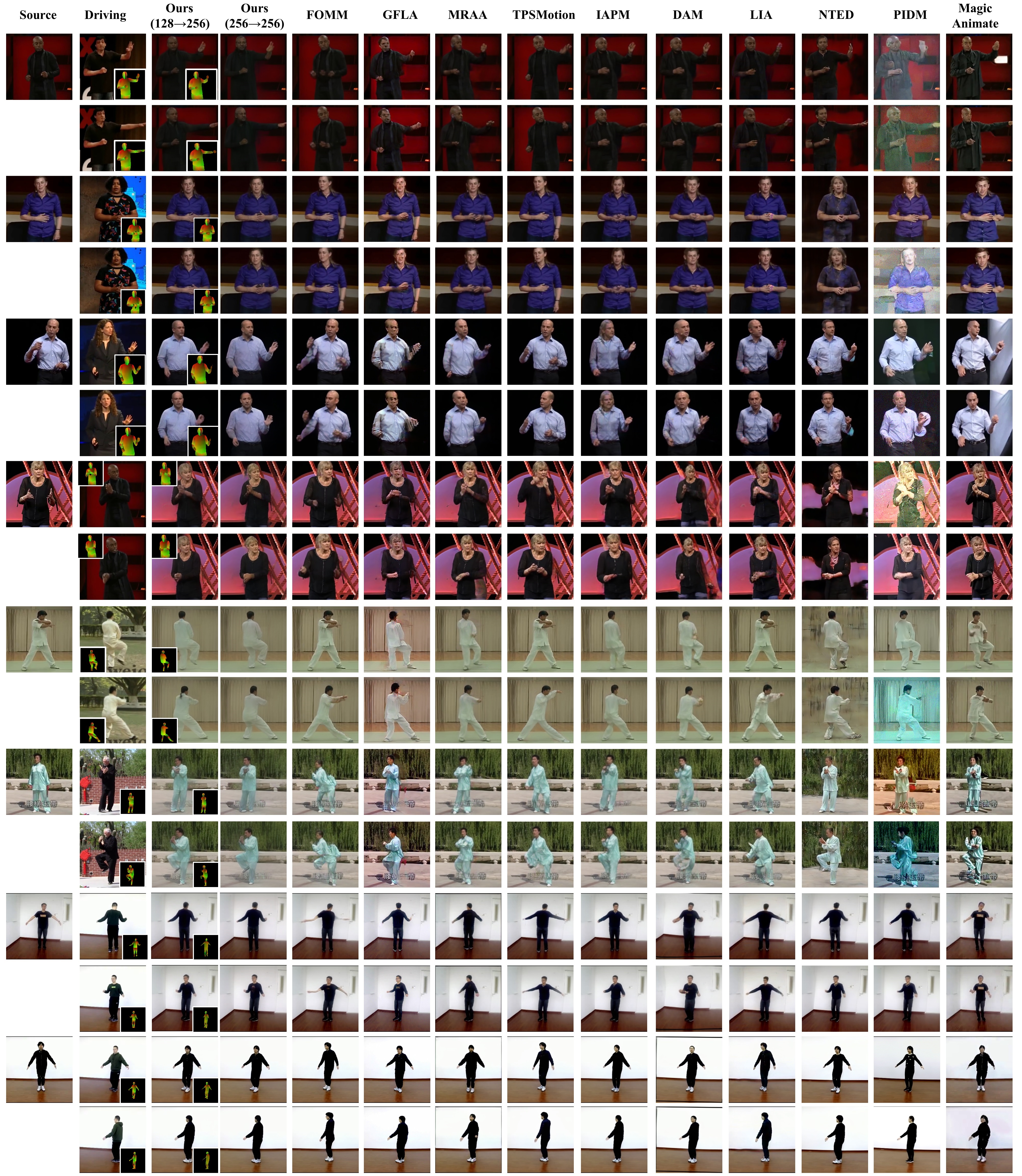}
  \caption{\textbf{Qualitative results in comparisons with state-of-the-art}. We show the results from TED-Talks (the first four sets), TaichiHD (the forth and fifth sets), and iPER dataset (the last two sets). It illustrates our model can animate both half and full human body images with (1) better geometry and details (see the first and second examples), (2) large variations in pose for front-to-back view and self-occlusion ({see the fifth and seventh examples}), (3) preserving the stability and consistency of animated video sequence (see the \href{https://github.com/AndrewChiyz/ORFPNT_supplementary/tree/main/suppl_video}{supplementary video} for more examples).}
  \label{fig2:cvanimation}
\end{figure*}

\textbf{2) Cross-Video Animation:} To evaluate the performance of cross-video animation, we randomly selected 50 source-driving video pairs from the test set of each benchmark. We fixed such source-driving video pairs across different methods. The first frame of a source video is regarded as the source image and will be animated by a driving video. Table \ref{tab:animation-sota} shows quantitative results in terms of AED, AKD and MKR in comparison with state-of-the-art. Since the ground truth for cross-video animation are not available, we cannot evaluate the L1 error under the animation settings. To assess identity preservation, we repeat the re-identification feature of the source frame as ground-truth $n$ times to obtain the AED (in which $n$ refers to the number of frames in a driving sequence). Moreover, to evaluate the quality of landmark alignment between an animated video and the driving video, the landmarks of the driving frames are predicted and treated as ground truth to calculate AKD and MKR. 

Quantitative results in Table \ref{tab:animation-sota} show that our model can achieve better performance on motion transfer, but not as well in identity preservation. The AKD scores of our model are lower than for other methods by a large margin across all datasets, with MKR also substantially better for TaiChiHD and iPER. {Although self-supervised methods, including FOMM, DAM, MRAA and TPSMotion achieved lower AED, the MKR and AKD values were higher than our proposed model. Moreover, according to our observation, FOMM and TPSMotion in multiple instances failed to significantly animate the source image, with facial and human body regions remaining mostly static. This lack of motion transfer may explain why AED can easily remain low for cross-identity animation. In comparison to DAM and MRAA, our model achieved competitive AED performance and broadly outperformed MRAA in terms of AKD as well as MKR (sans TED-Talks).}

{To further verify the scalability of the proposed model to large resolution, two sets of experiments on evaluation concerning the input and output resolutions with 256x256 have been performed. Specifically, we set the batch size to 8 and the number of training iterations to 600,000 during the training phase. For model structure, the super-resolution layer of the BlenderNet is removed. In addition, we have also withdrawn the multi-scale perceptual correctness loss and multi-scale affine transformation regularization terms in the loss function due to the high computational complexity and significant storage requirements. Moreover, to streamline the training process and reduce the time cost, our models are trained from scratch in an end-to-end manner. Quantitative results on three benchmark datasets, including TED-Talks, TaiChiHD, and iPER, are listed in Table \ref{tab:selfrec-sota} and Table \ref{tab:animation-sota} for self-reconstruction and cross-identity animation, respectively. It shows that the trained models, \ie, \textit{Ours (256$\rightarrow$256)} can achieve comparable performance by increasing the input and output resolution. The performance concerning AKD can be consistently improved across three benchmark datasets. However, the L1 distance and AED have not improved significantly. It indicates an upper bound of the performance may be limited by the default setting of the generator structure. It needs further investigation to introduce more powerful generators into our framework to improve video quality in future work.}

\textbf{3) Human evaluation:} We recruited 12 participants to score the quality of videos produced by different models. We randomly sampled 30 pairs of source and driving videos from the three datasets, along with the corresponding generated videos for different models. For each generated video, participants judged the quality based on the following four criteria: (1) \textit{Continuity (Cont.)}, which reflects the temporal fluency of the generated videos. (2) \textit{Appearance (App.)}, which considers whether the source identity is well preserved through the accurate recovery of facial, clothing, and hair textures. (3) \textit{Geometry (Geom.)}, which evaluates the correctness of the body pose. (4) \textit{Overall}, representing the overall rating score based on the aforementioned three aspects. Each criterion is scored on a scale of 1 to 5, in which ``1'' denotes the ``worst/unacceptable'' and ``5'' signifies the ``best performance''. Human evaluation results are listed in Table~\ref{tab:animation-sota}. It shows that our models achieve competitive scores across all aspects, and consistently outperform other methods, especially in the \textit{Geometry} and \textit{Overall} evaluations.

The qualitative results are visualized in Fig.~\ref{fig2:cvanimation}. We highlight our strengths for motion transfer in three aspects: 1) Recovering correct \emph{geometry} and details. For results from TED-Talks, the arm and hand pose of the source person are correctly transferred (see the first and second examples in Fig.~\ref{fig2:cvanimation}). Due to the semantic ambiguity of the unsupervised regions detected by the MRAA and FOMM models, the driving pose cannot be correctly transferred. 2) Handling back-to-front flips and \emph{heavy self-occlusion}. Our model can better handle front-to-back view motion transfer in the TaiChiHD dataset (see the fifth result in Fig.~\ref{fig2:cvanimation}), producing correct geometry despite substantial self-occlusion. 3) Preserving video animation stability and consistency. The examples from the iPER dataset (the last two results in Fig.~\ref{fig2:cvanimation}) show driving sequences with turning-around motion and large variation between source and driving images. It can be observed that our model generates a more stable video sequence in which the pose is correctly aligned with the driving video. Please see the \href{https://github.com/AndrewChiyz/ORFPNT_supplementary/tree/main/suppl_video}{supplementary video}.

\begin{table*}[htbp]
  \centering
  \caption{Ablation study on TED dataset.}
  \scalebox{1.0}{
    \begin{tabular}{l|ccccc|ccc}
      \hline
            & \multicolumn{5}{c|}{Self-reconstruction} & \multicolumn{3}{c}{Animation} \\
  \cline{2-9}          & L1↓   & FID↓  & AED↓  & MKR↓  & AKD↓  & AED↓  & MKR↓  & AKD↓ \\
  \hline
      \textit{2D baseline} & 0.0324 & 37.5687 & 0.1644 & 0.0090 & 2.2310 & 0.3557 & \cellcolor[rgb]{ .851,  .851,  .851}0.0173 & \cellcolor[rgb]{ .851,  .851,  .851}3.3832 \\
      \textit{2D baseline+BlenderNet} & 0.0363 & 30.7249 & 0.1727 & 0.0108 & 2.1405 & 0.3611 & 0.0212 & \cellcolor[rgb]{ .502,  .502,  .502}\textbf{3.1390} \\
      \textit{2D baseline+MotionNet} & \cellcolor[rgb]{ .851,  .851,  .851}0.0317 & 31.2702 & 0.1580 & \cellcolor[rgb]{ .651,  .651,  .651}0.0081 & 2.3928 & 0.3531 & 0.0223 & 4.2362 \\
      \textit{2D baseline+MotionNet+BlenderNet} & 0.0319 & \cellcolor[rgb]{ .651,  .651,  .651}\textbf{22.5662} & \cellcolor[rgb]{ .651,  .651,  .651}0.1486 & 0.0088 & \cellcolor[rgb]{ .502,  .502,  .502}1.9833 & \cellcolor[rgb]{ .502,  .502,  .502}\textbf{0.3481} & 0.0176 & \cellcolor[rgb]{ .651,  .651,  .651}3.3809 \\
      \textit{2D baseline+MotionNet+BlenderNet (w/o occl.map)} & 0.0321 & \cellcolor[rgb]{ .749,  .749,  .749}22.9708 & \cellcolor[rgb]{ .851,  .851,  .851}0.1487 & \cellcolor[rgb]{ .851,  .851,  .851}0.0086 & 2.3323 & 0.3508 & \cellcolor[rgb]{ .502,  .502,  .502}\textbf{0.0166} & 4.3218 \\
      \textit{2D baseline+MotionNet+BlenderNet (w SSFW)} & 0.0320 & \cellcolor[rgb]{ .851,  .851,  .851}23.1461 & 0.1540 & 0.0093 & \cellcolor[rgb]{ .851,  .851,  .851}2.1186 & \cellcolor[rgb]{ .502,  .502,  .502}\textbf{0.3481} & 0.0220 & 3.5512 \\
      \textit{2.5D baseline} & \cellcolor[rgb]{ .502,  .502,  .502}\textbf{0.0314} & 40.5923 & 0.2036 & 0.0116 & 3.4004 & 0.3827 & 0.0222 & 5.0950 \\
      \textit{2.5D baseline+BlenderNet} & 0.0349 & 35.8973 & 0.2163 & 0.0089 & 3.5312 & 0.4592 & \cellcolor[rgb]{ .651,  .651,  .651}0.0171 & 5.3624 \\
      \textit{2D+2.5D+MotionNet+BlenderNet(full)} & \cellcolor[rgb]{ .651,  .651,  .651}0.0316 & 26.3862 & \cellcolor[rgb]{ .502,  .502,  .502}\textbf{0.1478} & \cellcolor[rgb]{ .502,  .502,  .502}\textbf{0.0071} & \cellcolor[rgb]{ .651,  .651,  .651}2.1255 & \cellcolor[rgb]{ .651,  .651,  .651}0.3515 & 0.0186 & 3.6755 \\
      \hline
      \end{tabular}%
   }
  \label{tab:ablation-studies}%
\end{table*}%

\subsection{Ablation Studies}
To validate the effectiveness and contributions of our proposed modules, we run a number of ablations to analyze our model by incrementally introducing key modules or ideas to a baseline. The quantitative results of self-reconstruction and animation on the TED-Talks dataset are reported in Table \ref{tab:ablation-studies}. 
{Specifically, we trained a \textit{2D baseline} model by directly using the DensePose IUV map as input to investigate the influence of introducing DensePose annotation.} Under this setting, the RGB source image, and one-hot encoded source and driving DensePose with 24 human body parts (72-dims) are concatenated and directly fed into the generator for pose transfer. The baseline generator has the same architecture as ~\cite{cvprSiarohinWRCT21,nipsSiarohinLT0S19} without flow-based feature warping. The \textit{BlenderNet} has the same structure as in our full model and is adapted to synthesize high-resolution frames. {The \textit{2D+MotionNet} represents the model with our multi-scale feature warping idea in the 2D branch. The \textit{2D+MotionNet+BlenderNet} represents the model with the 2D branch and \textit{BlenderNet}. In addition, we further ablated the impact of multi-scale feature warping (\textit{MSFW}) and the occlusion map. Specifically, \textit{MSFW} is replaced by a single-scale feature warping (\textit{w SSFW}) setting at the largest flow $128\times128$, and the occlusion maps are removed (\textit{w/o occl. map}). For the \textit{2.5D baseline}, the encoder of \textit{MotionNet} is introduced to produce the condition signals of motion residuals for neural texture and DensePose translation. The \textit{2.5D baseline+BlenderNet} model consists of our proposed neural texture mapping network in the 2.5D branch, the encoder of \textit{MotionNet} and the \textit{BlenderNet}. The \textit{2D+2.5D+MotionNet+BlenderNet} refers to our full model.}

{It can be seen that the \textit{2D baseline} achieved good performance for self-reconstruction with lower L1 and AED, but relatively lower performance in animation. It can be regarded as a strong baseline and demonstrates the effectiveness of using DensePose as conditional supervision for motion transfer.} 

{The \textit{2D baseline+BlenderNet} achieved better performance in terms of AKD for animation but a slightly higher reconstruction loss for self-reconstruction in comparison with the \textit{2D baseline} model. In addition, the \textit{2D baseline+MotionNet} shows superior performance in comparison with the above two baseline models. It demonstrates the \textit{MotionNet} module can significantly improve the performance by introducing multi-scale feature warping. Moreover, the \textit{BlenderNet} module can further enhance the performance concerning AED and AKD for better pose transfer in comparison to other baseline models. Moreover, \textit{SSFW} degrades the motion transfer quality with larger MKR and AKD in comparison to the baseline (\textit{2D baseline+MotionNet+Blender}), while flow warping without occlusion map (\textit{w/o occl. map}) also degrades the performance, especially for AKD. It demonstrates the 2D branch with multi-scale feature warping preserves the appearance by leveraging the predicted dense motion flow. The reason for the higher AKD is related to the quality and stability of DensePose for calculating motion flow. The translated feature only with feature warping in the 2D branch may not be robust to the corrupted DensePose IUV map.}

{For ablation on the 2.5D branch, the \textit{2.5D baseline} achieved comparable performance in terms of L1 reconstruction loss but relatively poor performance in cross-identity animation. Moreover, the image quality can be degraded because of wrongly translated poses. We believe the reasons include: 1) Single branch training. The \textit{2.5D baseline} and \textit{2.5D baseline+BlenderNet} models with differentiable texture mapping may not benefit from single-stream training to render a high-quality appearance. From observation, the single 2.5D branch may not work well in transferring the source pose to a driving pose, especially for cross-identity pose transfer, 2) Model hyperparameter setting. For a fair comparison, we adopted the same network structure for the 2.5D branch as the full model. However, for the single 2.5D branch, the hyperparameter settings for modelling the diversity of motion change may be insufficient for texture atlas and DensePose translation. The model capability is limited by the length of latent vectors of $\alpha$ and $\rho$, the layers of the translation network, the length of neural texture feature vectors, etc.}

Our full model achieved better performance in comparison with other baseline models, demonstrating that the full model can take advantage of both 2D and 2.5D branches, by combining the features of flow warping and neural texture mapping to produce high-quality appearance and geometry.

\begin{figure*}
  \centering
   \includegraphics[width=0.92\linewidth]{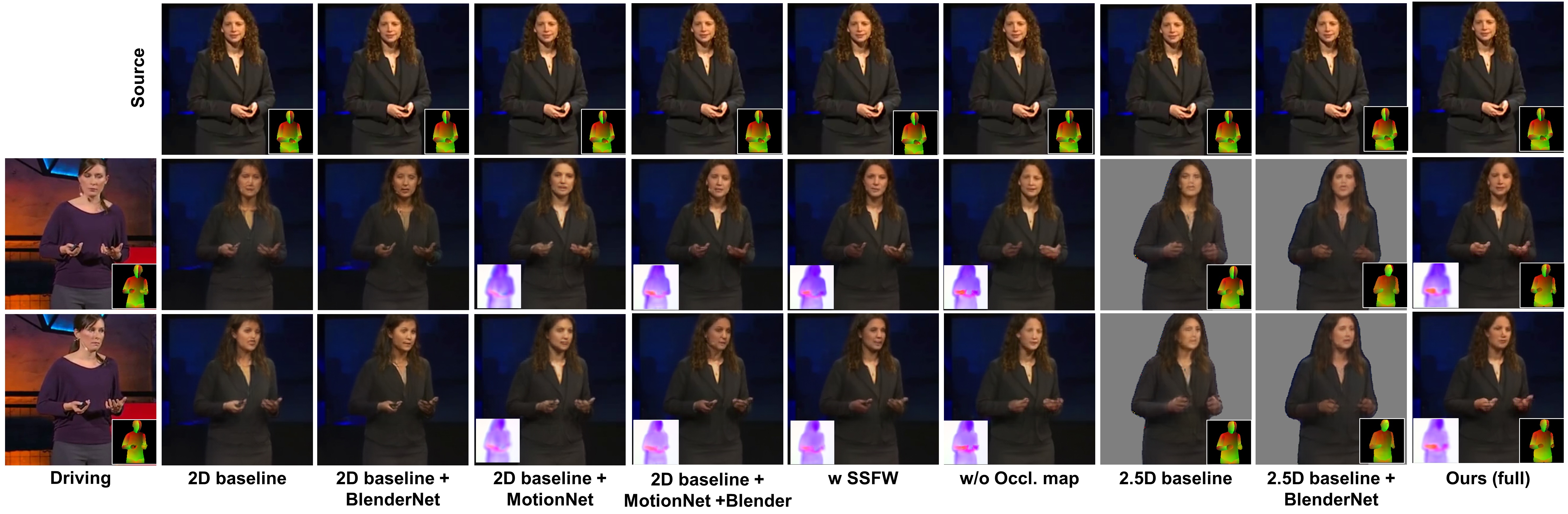}
  \caption{Qualitative comparisons for ablation study. We show the final generated images, dense motion flow and translated DensePose IUV map for animation.}
  \label{fig3:ablation}
\end{figure*}

We also visualize the animation results of different baseline models for ablation in Fig.~\ref{fig3:ablation}. {It shows that the \textit{2D baseline+MotionNet+BlenderNet} model can preserve more details of the source person, especially for the face. The \textit{2D baseline} and \textit{2D baseline+BlenderNet} models, without feature warping, produced blurry images, with facial, hair and clothing details lost. The \textit{2.5D baseline} and \textit{2.5D baseline+BlenderNet} models can translate the source body shape with correct geometry conditioning on the driving pose, but generate blurry images with low-quality appearance. Our full model can generate high-quality images with correct poses by jointly training both the 2D and 2.5D branches. The results in the last column of Fig.~\ref{fig3:ablation} show that images generated by the full model are more stable with better appearance and geometry.}

\subsection{Runtime Efficiency}
\label{sec:timeres}
{We evaluated the time cost and frames per second (FPS) on the TED dataset with a resolution of 256x256. We record the inference time for generating 1,000 frames and then average the total time to get the time cost for each frame and FPS. (For GFLA~\cite{tipRenLLL20}, we generate 1,002 frames due to the default setting of generating 6 frames each time). The time cost to generate keypoints (GFLA~\cite{tipRenLLL20}) and DensePose (ours) are not counted here. In our case, DensePose is the bottleneck at 20 FPS alone.}

\begin{table}[tb!]
  \centering
    \caption{Runtime cost and FPS on TED dataset.}
    \scalebox{0.92}{
    \begin{tabular}{lrrr}
    \hline
          & \multicolumn{1}{r}{\#params (M)} & \multicolumn{1}{l}{Time(s)} & \multicolumn{1}{l}{FPS} \\
    \hline
    FOMM  & 59.79 & 0.0165 & 60.59 \\
    {GFLA}  & 23.51 & 0.0185 & 54.02 \\
    MRAA  & \multicolumn{1}{l}{65.98(+AVD 2.29)} & 0.0138 & 72.32 \\
    {TPSMotion} & 85.10 & 0.0236 & 42.34 \\
    {IAPM}  & 89.46 & 0.0141 & 70.98 \\
    {DAM}   & 66.11 & 0.0136 & 73.27 \\
    {LIA} & 45.12 & {0.0339} & 29.50 \\
    {NTED} & 42.35  & 0.0187 & 53.48  \\
    {PIDM} & 180.36 & 4.6646  & 0.214  \\
    {MagicAnimate} & 2701.72
  & 0.4763  & 2.099 \\
    Ours  & 34.84 & 0.0170 & 58.68 \\
    \hline
    \end{tabular}%
  }
  \label{tab:runtime}%
\end{table}%

\section{Discussion and Limitations}\label{sec:discusion}
\subsection{Incorrect DensePose Annotation.} 
\begin{figure}[tb!]
  \centering
   \includegraphics[width=0.9\linewidth]{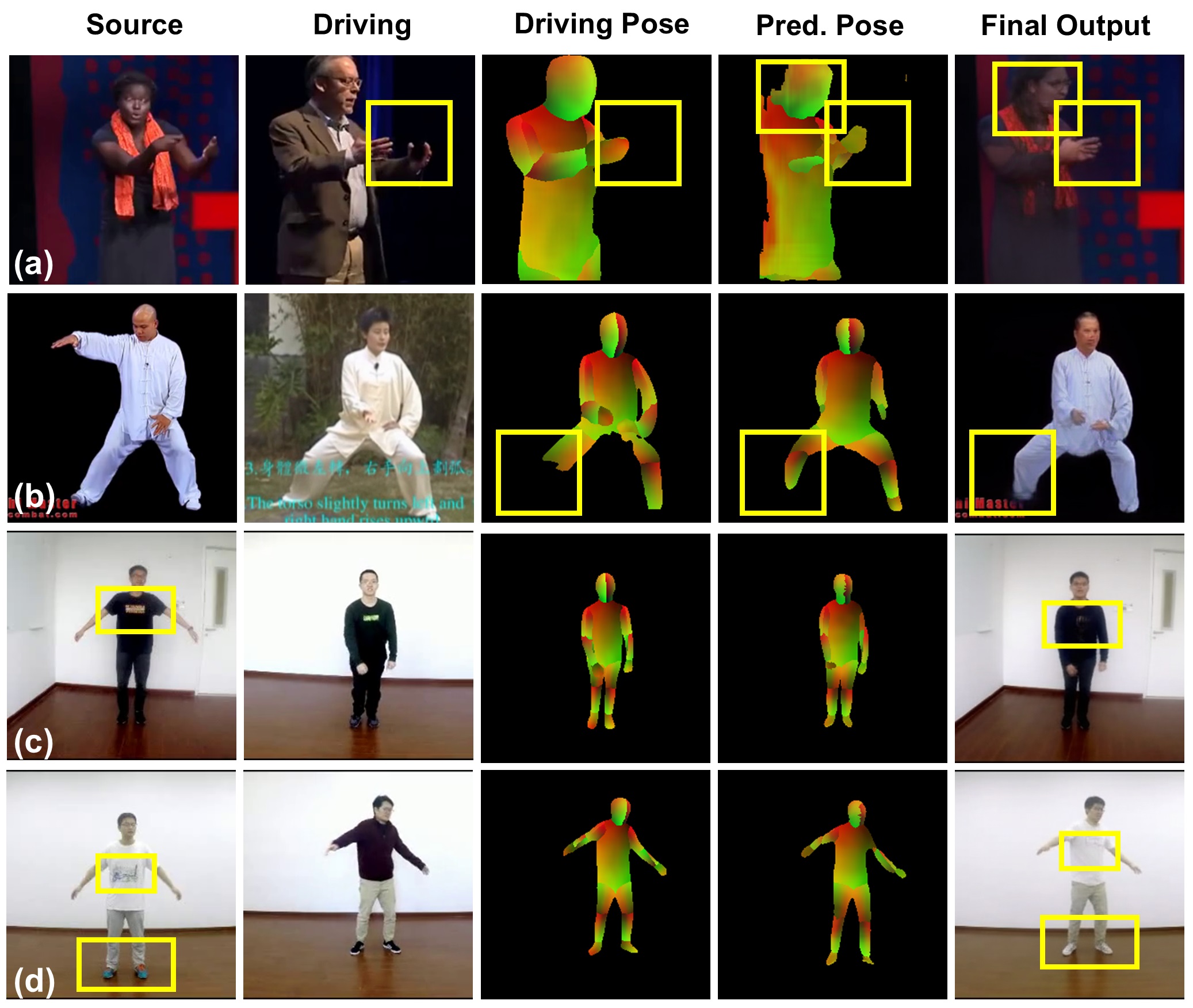}
  \caption{Failure cases produced by our model with (1) broken DensePose IUV maps (as shown in (a) and (b)), (2) inaccurate fine-grained appearance (depicted in (c) and (d)).}
  \label{fig4:failurecases}
\end{figure}

Due to the proposed neural texture mapping and motion flow feature warping branches both relying on the stability of DensePose, the quality of the pseudo-ground-truth DensePose annotation can largely impact the performance of our model. According to our observation, the stability and quality of motion transfer can be degraded by the incorrect DensePose IUV map, including the IUV map with dark holes, missing human body parts, etc. It can lead to obvious artefacts in the missing regions for video synthesis. Our proposed 2.5D branch for DensePose Transfer Network can alleviate such problems and stabilize the generated videos by completing the driving IUV map, but the model will fail to correct geometry errors when the correspondence in the IUV map is severely missing, such as arms, hands or feet may be missing~(see Fig.~\ref{fig4:failurecases} (a) and (b)). In addition, the DensePose annotations can also affect the dense motion flow estimation in the 2D branch of our model. The missing parts may lead to abnormal values in the displacement vector. The feature could be wrongly warped.

{\subsection{Fine-grained motion transfer and appearance recovering} 
Because the DensePose is based on the SMPL model, our proposed model cannot capture subtle motion in the facial part, and thus cannot achieve facial expression and lips motion transfer. Therefore, further investigation is required to generate fine-grained dense motion flow for subtle motion transfer. It would be interesting to adopt some parametric 3D models with body pose, hand pose, and facial expression to produce dense motion flow conditional on the relevant predicted parameters. In addition, our model cannot preserve the body shape of the source person well in motion transfer. We believe it is related to the robustness of DensePose transfer and the strategy of multi-modal fusion. Moreover, according to our observation, the model may fail to render a faithful and consistent appearance in fine-grained human body parts, \eg, face, hands, shoes and clothing logos (see Fig.~\ref{fig4:failurecases} (c) and (d)). It may be attributed to the information loss in the encoding and decoding process and the limited information of the source person under the one-shot setting. Similar artefacts can also be observed from the results generated by other SOTA models. In practice, generating such fine-grained regions in fully-body motion transfer without hallucination remains particularly challenging in the one-shot setting. We believe it is promising to extend our model for person-specific or few-shot human motion transfer tasks. In future, it is worth further investigation to apply our model to recover fine-grained appearance by predicting the complete texture atlas of a specific person.}

\subsection{Generalization ability on high-speed motion and other benchmarks.} For videos with high-speed motion, if a single frame contains strong motion blur or severe occlusion in a short frame, it is universally challenging for most video analysis methods. The current DensePose methods cannot handle such situations, which will also degrade the final results of our model. For generalization ability on other benchmarks, e.g., VoxCeleb~\cite{nagrani2020}, MGIFs~\cite{mgifsSiarohinLT0S19}, since our model is mainly focused on the human body motion transfer, it is potential to apply our model to animate other objects if the UV map or other type of geometric representation could be obtained. Specifically, for VoxCeleb~\cite{nagrani2020}, which is an audio-visual dataset and mainly explored for face-related synthesis, our model can be applied to animate faces if higher-resolution UV maps can be obtained. However, those datasets typically do not have complex self-occlusion cases, unlike bodies, which is the problem we want to address by considering the complexity of human body motion transfer.

\section{Conclusion}
\label{sec:conclusion}
Appearance and geometry are critical for high-quality pose-guided human image animation. In this paper, we present a unified human motion transfer framework to combine feature warping and neural texture mapping for better recovering the appearance and the driving pose. We also propose to implicitly decouple the neural texture atlas translation and DensePose translation into a two-branch pipeline for correcting the geometric errors in pseudo-ground-truth DensePose annotation. Experimental results demonstrate our model benefits from both appearance and geometry branches by multi-modal feature fusion and joint training. Qualitative results show that our model can work well for challenging motion patterns with large variations in appearance and pose, \eg, turning-around, self-occlusion, and front-to-back view motion transfer in comparison with SOTA models. In future, it would be interesting to investigate the controllable human body image rendering by manipulating different human body parts. In addition, subtle motions, including eyes, lips and facial expressions, remain challenging and need to be further explored.

\section*{acknowledgements}
This work is supported in part by the National Natural Science Foundation of China (NSFC) under grants: 62302104, the Natural Science Foundation of Guangdong Province under grants: 2023A1515012884, the Science and Technology Program of Guangzhou under grant SL2023A04J01625. This study is also supported under the RIE2020 Industry Alignment Fund – Industry Collaboration Projects (IAF-ICP) Funding Initiative, as well as cash and in-kind contribution from Singapore Telecommunications Limited (Singtel), through Singtel Cognitive and Artificial Intelligence Lab for Enterprises (SCALE@NTU). 


\bibliographystyle{IEEEtran}
\bibliography{ref_rec}

\begin{thebibliography}{10}
\providecommand{\url}[1]{#1}
\csname url@samestyle\endcsname
\providecommand{\newblock}{\relax}
\providecommand{\bibinfo}[2]{#2}
\providecommand{\BIBentrySTDinterwordspacing}{\spaceskip=0pt\relax}
\providecommand{\BIBentryALTinterwordstretchfactor}{4}
\providecommand{\BIBentryALTinterwordspacing}{\spaceskip=\fontdimen2\font plus
\BIBentryALTinterwordstretchfactor\fontdimen3\font minus \fontdimen4\font\relax}
\providecommand{\BIBforeignlanguage}[2]{{%
\expandafter\ifx\csname l@#1\endcsname\relax
\typeout{** WARNING: IEEEtran.bst: No hyphenation pattern has been}%
\typeout{** loaded for the language `#1'. Using the pattern for}%
\typeout{** the default language instead.}%
\else
\language=\csname l@#1\endcsname
\fi
#2}}
\providecommand{\BIBdecl}{\relax}
\BIBdecl

\bibitem{liu2019neural}
L.~Liu, W.~Xu, M.~Zollhoefer \emph{et~al.}, ``Neural rendering and reenactment of human actor videos,'' \emph{ACM TOG}, vol.~38, no.~5, pp. 1--14, 2019.

\bibitem{cgfTewariFTSLSMSSN20}
A.~Tewari, O.~Fried, J.~Thies \emph{et~al.}, ``State of the art on neural rendering,'' \emph{Comput. Graph. Forum}, vol.~39, no.~2, pp. 701--727, 2020.

\bibitem{cvprSiarohinSLS18}
A.~Siarohin, E.~Sangineto, S.~Lathuili{\`{e}}re \emph{et~al.}, ``Deformable gans for pose-based human image generation,'' in \emph{CVPR}, 2018, pp. 3408--3416.

\bibitem{nipsSiarohinLT0S19}
A.~Siarohin, S.~Lathuili{\`{e}}re, S.~Tulyakov \emph{et~al.}, ``First order motion model for image animation,'' in \emph{NIPS}, 2019, pp. 7135--7145.

\bibitem{cvprSiarohinWRCT21}
A.~Siarohin, O.~J. Woodford, J.~Ren \emph{et~al.}, ``Motion representations for articulated animation,'' in \emph{CVPR}, 2021, pp. 13\,653--13\,662.

\bibitem{TaoG0D23flow}
J.~Tao, S.~Gu, W.~Li \emph{et~al.}, ``Learning motion refinement for unsupervised face animation,'' in \emph{NeurIPS}, 2023.

\bibitem{cvprZhuHSYWB19}
Z.~Zhu, T.~Huang, B.~Shi \emph{et~al.}, ``Progressive pose attention transfer for person image generation,'' in \emph{{CVPR}}, 2019, pp. 2347--2356.

\bibitem{RenF0LLcvpr22}
Y.~Ren, X.~Fan, G.~Li \emph{et~al.}, ``Neural texture extraction and distribution for controllable person image synthesis,'' in \emph{CVPR}, 2022, pp. 13\,525--13\,534.

\bibitem{cvprBhuniaKCALSK23}
A.~K. Bhunia, S.~H. Khan, H.~Cholakkal \emph{et~al.}, ``Person image synthesis via denoising diffusion model,'' in \emph{{CVPR}}, 2023, pp. 5968--5976.

\bibitem{Zhang_2020_CVPR}
P.~Zhang, B.~Zhang, D.~Chen \emph{et~al.}, ``Cross-domain correspondence learning for exemplar-based image translation,'' in \emph{CVPR}, 2020, pp. 5143--5153.

\bibitem{Zhou_2021_CVPR}
X.~Zhou, B.~Zhang, T.~Zhang \emph{et~al.}, ``Cocosnet v2: Full-resolution correspondence learning for image translation,'' in \emph{CVPR}, 2021, pp. 11\,465--11\,475.

\bibitem{nipsWang0ZYTKC18}
T.~Wang, M.~Liu, J.~Zhu \emph{et~al.}, ``Video-to-video synthesis,'' in \emph{NeurIPS}, 2018, pp. 1152--1164.

\bibitem{iccvChanGZE19}
C.~Chan, S.~Ginosar, T.~Zhou \emph{et~al.}, ``Everybody dance now,'' in \emph{{ICCV}}, 2019, pp. 5932--5941.

\bibitem{nipsWang0TLCK19}
T.~Wang, M.~Liu, A.~Tao \emph{et~al.}, ``Few-shot video-to-video synthesis,'' in \emph{NeurIPS}, 2019, pp. 5014--5025.

\bibitem{SunHWLLGpami23}
Y.~Sun, H.~Huang, X.~Wang \emph{et~al.}, ``Robust pose transfer with dynamic details using neural video rendering,'' \emph{{IEEE} Trans. Pattern Anal. Mach. Intell.}, vol.~45, no.~2, pp. 2660--2666, 2023.

\bibitem{eccvNeverovaGK18}
N.~Neverova, R.~A. G{\"{u}}ler, and I.~Kokkinos, ``Dense pose transfer,'' in \emph{ECCV}, vol. 11207, 2018, pp. 128--143.

\bibitem{eccvSarkarMXGT20}
K.~Sarkar, D.~Mehta, W.~Xu \emph{et~al.}, ``Neural re-rendering of humans from a single image,'' in \emph{{ECCV}}, vol. 12356, 2020, pp. 596--613.

\bibitem{huang2021few}
Z.~Huang, X.~Han, J.~Xu \emph{et~al.}, ``Few-shot human motion transfer by personalized geometry and texture modeling,'' in \emph{CVPR}, 2021, pp. 2297--2306.

\bibitem{tcsvtYangYZJZS22}
C.~Yang, S.~Yao, Z.~Zhou \emph{et~al.}, ``Poxture: Human posture imitation using neural texture,'' \emph{{IEEE} Trans. Circuits Syst. Video Technol.}, vol.~32, no.~12, pp. 8537--8549, 2022.

\bibitem{tmmLiuZCZ22}
J.~Liu, Y.~Zhao, S.~Chen \emph{et~al.}, ``A 3d mesh-based lifting-and-projection network for human pose transfer,'' \emph{{IEEE} Trans. Multim.}, vol.~24, pp. 4314--4327, 2022.

\bibitem{corrdreampose23}
J.~Karras, A.~Holynski, T.~Wang \emph{et~al.}, ``Dreampose: Fashion image-to-video synthesis via stable diffusion,'' \emph{CoRR}, vol. abs/2304.06025, 2023.

\bibitem{cvprGulerNK18}
R.~A. G{\"{u}}ler, N.~Neverova, and I.~Kokkinos, ``Densepose: Dense human pose estimation in the wild,'' in \emph{CVPR}, 2018, pp. 7297--7306.

\bibitem{togLoperM0PB15}
M.~Loper, N.~Mahmood, J.~Romero \emph{et~al.}, ``{SMPL:} a skinned multi-person linear model,'' \emph{{ACM} Trans. Graph.}, vol.~34, no.~6, pp. 248:1--248:16, 2015.

\bibitem{tcybXiaMLZ23}
G.~Xia, F.~Ma, Q.~Liu \emph{et~al.}, ``Pose-driven realistic 2-d motion synthesis,'' \emph{{IEEE} Trans. Cybern.}, vol.~53, no.~4, pp. 2412--2425, 2023.

\bibitem{KappelGEHSCTM21}
M.~Kappel, V.~Golyanik, M.~Elgharib \emph{et~al.}, ``High-fidelity neural human motion transfer from monocular video,'' in \emph{CVPR}, 2021, pp. 1541--1550.

\bibitem{tipAlbarracinR22}
J.~F.~H. Albarrac{\'{\i}}n and A.~R. Rivera, ``Video reenactment as inductive bias for content-motion disentanglement,'' \emph{{IEEE} Trans. Image Process.}, vol.~31, pp. 2365--2374, 2022.

\bibitem{tmmWeiXSH21}
D.~Wei, X.~Xu, H.~Shen \emph{et~al.}, ``{GAC-GAN:} {A} general method for appearance-controllable human video motion transfer,'' \emph{{IEEE} Trans. Multim.}, vol.~23, pp. 2457--2470, 2021.

\bibitem{cvprZhang0L021}
J.~Zhang, K.~Li, Y.~Lai \emph{et~al.}, ``{PISE:} person image synthesis and editing with decoupled {GAN},'' in \emph{{CVPR}}, 2021, pp. 7982--7990.

\bibitem{nipsDongLGLZY18}
H.~Dong, X.~Liang, K.~Gong \emph{et~al.}, ``Soft-gated warping-gan for pose-guided person image synthesis,'' in \emph{NIPS}, 2018, pp. 472--482.

\bibitem{tipRenLLL20}
Y.~Ren, G.~Li, S.~Liu \emph{et~al.}, ``Deep spatial transformation for pose-guided person image generation and animation,'' \emph{{IEEE} Trans. Image Process.}, vol.~29, pp. 8622--8635, 2020.

\bibitem{TaoWGJLD22}
J.~Tao, B.~Wang, T.~Ge \emph{et~al.}, ``Motion transformer for unsupervised image animation,'' in \emph{ECCV}, vol. 13676, 2022, pp. 702--719.

\bibitem{cvprMenMJML20}
Y.~Men, Y.~Mao, Y.~Jiang \emph{et~al.}, ``Controllable person image synthesis with attribute-decomposed {GAN},'' in \emph{CVPR}, 2020, pp. 5083--5092.

\bibitem{aaaiWeiXSH21}
D.~Wei, X.~Xu, H.~Shen \emph{et~al.}, ``{C2F-FWN:} coarse-to-fine flow warping network for spatial-temporal consistent motion transfer,'' in \emph{AAAI}, 2021, pp. 2852--2860.

\bibitem{YangLLXGY022}
Q.~Yang, X.~Liu, W.~Liu \emph{et~al.}, ``{REMOT:} {A} region-to-whole framework for realistic human motion transfer,'' in \emph{ACM MM}, 2022, pp. 1128--1137.

\bibitem{yoon2021pose}
J.~S. Yoon, L.~Liu, V.~Golyanik \emph{et~al.}, ``Pose-guided human animation from a single image in the wild,'' in \emph{CVPR}, 2021, pp. 15\,039--15\,048.

\bibitem{bmvcZablotskaiaSZS19}
P.~Zablotskaia, A.~Siarohin, B.~Zhao \emph{et~al.}, ``Dwnet: Dense warp-based network for pose-guided human video generation,'' in \emph{{BMVC}}, 2019, p.~51.

\bibitem{3dimarkarLGT21}
K.~Sarkar, L.~Liu, V.~Golyanik \emph{et~al.}, ``Humangan: {A} generative model of human images,'' in \emph{3DV}, 2021, pp. 258--267.

\bibitem{cvprLiHL19}
Y.~Li, C.~Huang, and C.~C. Loy, ``Dense intrinsic appearance flow for human pose transfer,'' in \emph{CVPR}, 2019, pp. 3693--3702.

\bibitem{iccvLiuPML0G19}
W.~Liu, Z.~Piao, J.~Min \emph{et~al.}, ``Liquid warping {GAN:} {A} unified framework for human motion imitation, appearance transfer and novel view synthesis,'' in \emph{ICCV}, 2019, pp. 5903--5912.

\bibitem{cvprHuangHXZ21}
Z.~Huang, X.~Han, J.~Xu \emph{et~al.}, ``Few-shot human motion transfer by personalized geometry and texture modeling,'' in \emph{{CVPR}}, 2021, pp. 2297--2306.

\bibitem{thies2019deferred}
J.~Thies, M.~Zollh{\"o}fer, and M.~Nie{\ss}ner, ``Deferred neural rendering: Image synthesis using neural textures,'' \emph{ACM TOG}, vol.~38, no.~4, pp. 1--12, 2019.

\bibitem{nipsHoJA20}
J.~Ho, A.~Jain, and P.~Abbeel, ``Denoising diffusion probabilistic models,'' in \emph{NeurIPS}, H.~Larochelle, M.~Ranzato, R.~Hadsell \emph{et~al.}, Eds., vol.~33, 2020, pp. 6840--6851.

\bibitem{icmlRadfordKHRGASAM21}
A.~Radford, J.~W. Kim, C.~Hallacy \emph{et~al.}, ``Learning transferable visual models from natural language supervision,'' in \emph{{ICML}}, vol. 139, 2021, pp. 8748--8763.

\bibitem{corrabs-2311-17117}
L.~Hu, X.~Gao, P.~Zhang \emph{et~al.}, ``Animate anyone: Consistent and controllable image-to-video synthesis for character animation,'' \emph{CoRR}, vol. abs/2311.17117, 2023.

\bibitem{iccv23zhang2023adding}
L.~Zhang and M.~Agrawala, ``Adding conditional control to text-to-image diffusion models,'' in \emph{ICCV}, 2023, pp. 3836--3847.

\bibitem{corranimatediff}
Y.~Guo, C.~Yang, A.~Rao \emph{et~al.}, ``Animatediff: Animate your personalized text-to-image diffusion models without specific tuning,'' \emph{CoRR}, vol. abs/2307.04725, 2023.

\bibitem{corrmagicanimate}
Z.~Xu, J.~Zhang, J.~H. Liew \emph{et~al.}, ``Magicanimate: Temporally consistent human image animation using diffusion model,'' in \emph{CVPR}, 2024, pp. 1481--1490.

\bibitem{cvprNiSLHM23}
H.~Ni, C.~Shi, K.~Li \emph{et~al.}, ``Conditional image-to-video generation with latent flow diffusion models,'' in \emph{{CVPR}}, 2023, pp. 18\,444--18\,455.

\bibitem{naaclDevlinCLT19}
J.~Devlin, M.~Chang, K.~Lee \emph{et~al.}, ``{BERT:} pre-training of deep bidirectional transformers for language understanding,'' in \emph{{NAACL-HLT}}, 2019, pp. 4171--4186.

\bibitem{eccvwRenCTFSY20}
J.~Ren, M.~Chai, S.~Tulyakov \emph{et~al.}, ``Human motion transfer from poses in the wild,'' in \emph{{ECCV} Workshops}, vol. 12537, 2020, pp. 262--279.

\bibitem{cvprIlgMSKDB17}
E.~Ilg, N.~Mayer, T.~Saikia \emph{et~al.}, ``Flownet 2.0: Evolution of optical flow estimation with deep networks,'' in \emph{CVPR}, 2017, pp. 1647--1655.

\bibitem{cvprRajTHVSL21}
A.~Raj, J.~Tanke, J.~Hays \emph{et~al.}, ``{ANR:} articulated neural rendering for virtual avatars,'' in \emph{{CVPR}}, 2021, pp. 3722--3731.

\bibitem{iccvMaoLXLWS17}
X.~Mao, Q.~Li, H.~Xie \emph{et~al.}, ``Least squares generative adversarial networks,'' in \emph{{ICCV}}, 2017, pp. 2813--2821.

\bibitem{cvprHeZRS16}
K.~He, X.~Zhang, S.~Ren \emph{et~al.}, ``Deep residual learning for image recognition,'' in \emph{{CVPR}}, 2016, pp. 770--778.

\bibitem{corrChenPSA17}
L.~Chen, G.~Papandreou, F.~Schroff \emph{et~al.}, ``Rethinking atrous convolution for semantic image segmentation,'' \emph{CoRR}, vol. abs/1706.05587, 2017.

\bibitem{wacvLinYSS22}
S.~Lin, L.~Yang, I.~Saleemi \emph{et~al.}, ``Robust high-resolution video matting with temporal guidance,'' in \emph{{WACV}}, 2022, pp. 3132--3141.

\bibitem{fidHeuselRUNH17}
M.~Heusel, H.~Ramsauer, T.~Unterthiner \emph{et~al.}, ``Gans trained by a two time-scale update rule converge to a local nash equilibrium,'' in \emph{NeurIPS}, 2017, pp. 6626--6637.

\bibitem{corrHermansBL17}
A.~Hermans, L.~Beyer, and B.~Leibe, ``In defense of the triplet loss for person re-identification,'' \emph{CoRR}, vol. abs/1703.07737, 2017.

\bibitem{cvprCaoSWS17}
Z.~Cao, T.~Simon, S.~Wei \emph{et~al.}, ``Realtime multi-person 2d pose estimation using part affinity fields,'' in \emph{CVPR}, 2017, pp. 1302--1310.

\bibitem{corrKingmaB14}
D.~P. Kingma and J.~Ba, ``Adam: {A} method for stochastic optimization,'' in \emph{{ICLR}}, 2015.

\bibitem{TaoWXGJLD22}
J.~Tao, B.~Wang, B.~Xu \emph{et~al.}, ``Structure-aware motion transfer with deformable anchor model,'' in \emph{CVPR}, 2022, pp. 3627--3636.

\bibitem{cvprShalevW22}
Y.~Shalev and L.~Wolf, ``Image animation with perturbed masks,'' in \emph{CVPR}, 2022, pp. 3637--3646.

\bibitem{cvprZhaoZ22}
J.~Zhao and H.~Zhang, ``Thin-plate spline motion model for image animation,'' in \emph{{CVPR}}, 2022, pp. 3647--3656.

\bibitem{iclrWangYBD22}
Y.~Wang, D.~Yang, F.~Br{\'{e}}mond \emph{et~al.}, ``Latent image animator: Learning to animate images via latent space navigation,'' in \emph{ICLR}, 2022.

\bibitem{nagrani2020}
A.~Nagrani, J.~S. Chung, W.~Xie \emph{et~al.}, ``Voxceleb: Large-scale speaker verification in the wild,'' \emph{Comput. Speech Lang.}, vol.~60, 2020.

\bibitem{mgifsSiarohinLT0S19}
A.~Siarohin, S.~Lathuili{\`{e}}re, S.~Tulyakov \emph{et~al.}, ``Animating arbitrary objects via deep motion transfer,'' in \emph{{CVPR}}, 2019, pp. 2377--2386.

\end{thebibliography}

%

\begin{IEEEbiography}[{\includegraphics[width=1in,height=1.25in,keepaspectratio]{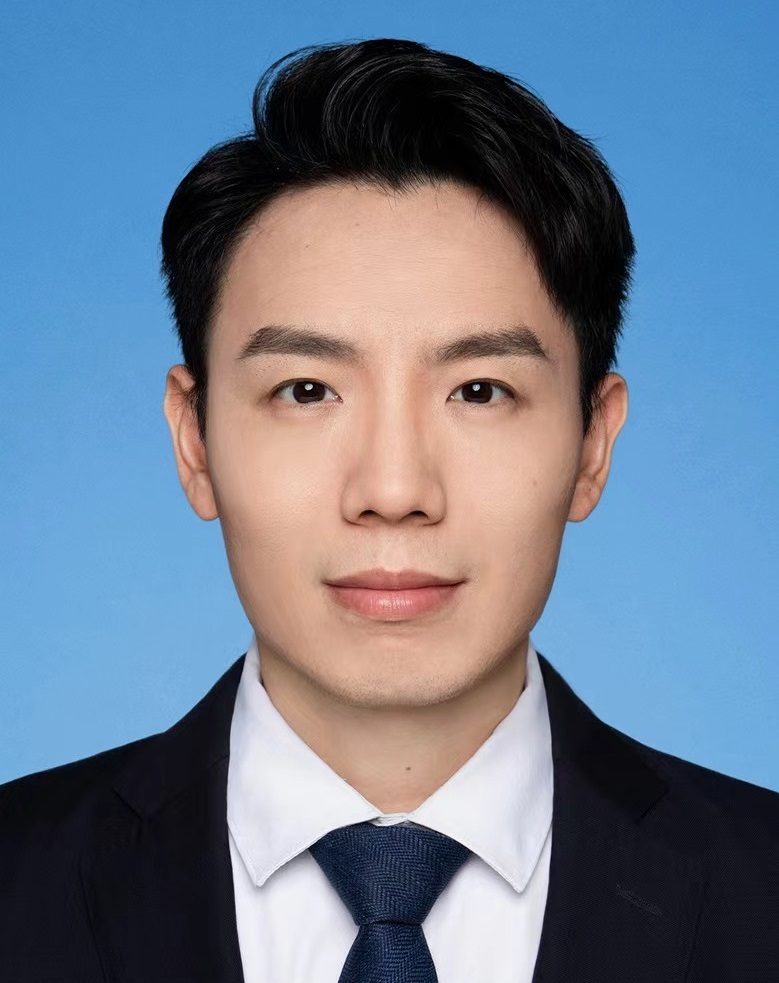}}]{Yuzhu Ji} (Member, IEEE) received a B.S. degree in computer science from the PLA Information Engineering University, Zhengzhou, China, in 2012, and the M.S. and Ph.D. degrees from the Department of Computer Science, Harbin Institute of Technology Shenzhen, China, in 2015 and 2019. He was a Post-Doctoral Research Fellow with the School of Computer Science and Engineering, Nanyang Technological University, from 2020 to 2022. He is currently an Associate Professor at the School of Computer Science and Technology, Guangdong University of Technology, Guangzhou, China. His current research interests include video analysis and synthesis, human motion transfer, and image segmentation-related topics in computer vision. He has served as a reviewer of several top conferences and journals, including CVPR, ICCV, ECCV, ACM MM, T-MM and TNNLS.
\end{IEEEbiography}

\begin{IEEEbiography}[{\includegraphics[width=1in,height=1.25in,keepaspectratio]{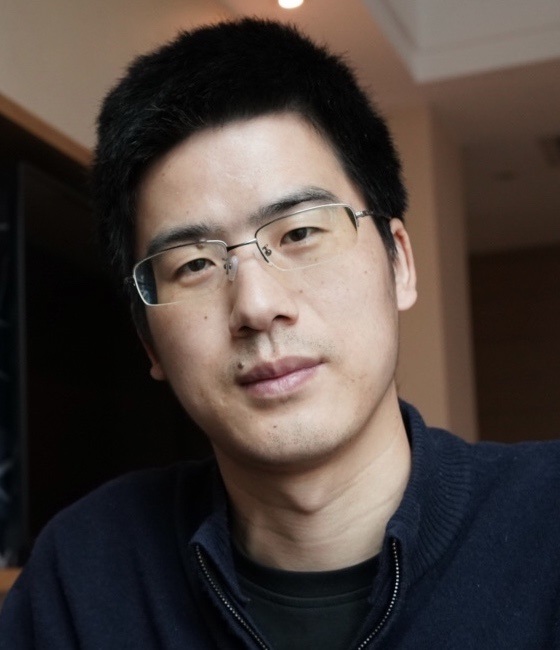}}]{Chuanxia Zheng} (Member, IEEE) is currently a \emph{Marie Skłodowska-Curie Actions (MSCA) Fellow} in VGG at the University of Oxford and \emph{DAAD AInet fellow} of Germany.
  He received his PhD in 2021 from Nanyang Technological University, Singapore,
  and obtained the \emph{NTU Outstanding PhD thesis award}.
  Chuanxia's research interests are broadly in computer vision and machine learning, especially for Generative AI in multi-modality (1D, 2D, 3D, and 4D) generation.
  He is author of more than 30 peer-reviewed publications in the top machine vision and artificial intelligence conferences and journals.
  His current and past research services include serving as Area Chair for ACM Multimedia, BMVC, organizing the Workshop for CVPR, and being the reviewer for TPAMI, IJCV, IEEE T-IP, T-MM, and CVPR, ICCV, ECCV, NeurIPS, ICML, ICLR, SIGGRAPH.
\end{IEEEbiography}

\begin{IEEEbiography}[{\includegraphics[width=1in,height=1.25in,keepaspectratio]{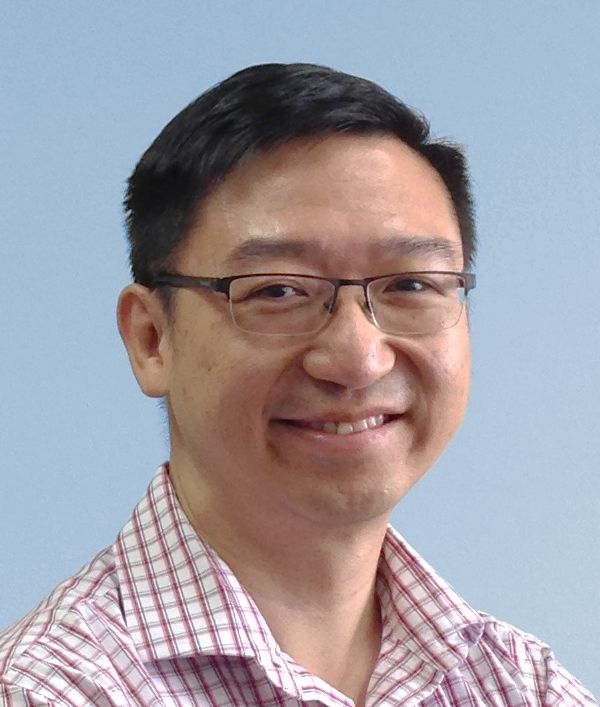}}]{Tat-Jen Cham} is a Professor of Computer Science in the College of Computing and Data Science, Nanyang Technological University, Singapore. He received his BA in Engineering in 1993 and his PhD in 1996, both from the University of Cambridge. He was previously a Jesus College Research Fellow in Science (1996-97), and a research scientist at DEC/Compaq Research Labs in Cambridge, MA, USA (1998-2021). Tat-Jen has received multiple paper awards, including the best paper at ECCV'96, and is an inventor on eight patents. Tat-Jen has been the principal investigator on projects that include those based in the Rehabilitation Research Institute of Singapore (RRIS), the Singtel Cognitive \& AI Lab (SCALE@NTU), Singapore-ETH Centre's Future Cities Lab, and the NRF BeingThere / BeingTogether Centres on 3D Telepresence. His current and past research services include being an Associate Editor for IEEE T-MM, CVIU and IJCV, as well as an Area Chair for CVPR, ICCV, ECCV and NeurIPS conferences. He was also a General Chair for ACCV 2014. Tat-Jen's research interests are broadly in computer vision and machine learning, with a focus on deep learning generative methods that can exploit semantic and contextual cues, for applications such as 3D telepresence and metaverses.
\end{IEEEbiography}

\end{document}